\documentclass[10pt,journal,compsoc]{IEEEtran}
\usepackage{bbm}
\usepackage{amsmath}
\usepackage{epsfig}
\usepackage{graphicx}
\usepackage{amssymb}
\usepackage{enumitem}
\usepackage{endnotes}
\usepackage{tabularx}
\newcolumntype{Y}{>{\centering\arraybackslash}X}
\usepackage{makecell}
\usepackage{multirow}
\usepackage{mathtools}
\usepackage{booktabs}
\usepackage[symbol]{footmisc}
\usepackage{xcolor}
\usepackage{subfigure}
\usepackage{color, soul}
\usepackage[caption=false,font=footnotesize]{subfig}
\usepackage[ruled, vlined, linesnumbered]{algorithm2e}

\ifCLASSOPTIONcompsoc
  \usepackage[nocompress]{cite}
\else
  \usepackage{cite}
\fi
\ifCLASSINFOpdf
\else
\fi
\begin{document}
%
\title{Face Shape-Guided Deep Feature Alignment for\\Face Recognition Robust to Face Misalignment}
%
%
%
%

\author{Hyung-Il Kim,~\IEEEmembership{Member,~IEEE,} Kimin Yun, and Yong Man Ro,~\IEEEmembership{Senior Member,~IEEE,}
\IEEEcompsocitemizethanks{
\IEEEcompsocthanksitem H.-I. Kim and K. Yun are with Visual Intelligence Research Section, Artificial Intelligence Research Laboratory, Electronics and Telecommunications Research Institute (ETRI), Daejeon, 34129, South Korea (e-mail: \{hikim, kimin.yun\}@etri.re.kr)
\IEEEcompsocthanksitem Y. M. Ro is with Image and Video Systems Lab, School of Electrical Engineering, Korea Advanced Institute of Science and Technology (KAIST), Daejeon, 34141, South Korea (e-mail: ymro@kaist.ac.kr)
}
\thanks{Corresponding Author: Yong Man Ro (email: ymro@kaist.ac.kr)}
}

%
%

\markboth{}%
{Shell \MakeLowercase{\textit{et al.}}: Bare Demo of IEEEtran.cls for Biometrics Council Journals}
\makeatletter
\long\def\@IEEEtitleabstractindextextbox#1{\parbox{0.922\textwidth}{#1}}
\makeatother
\IEEEtitleabstractindextext{%
\begin{abstract}
For the past decades, face recognition (FR) has been actively studied in computer vision and pattern recognition society. Recently, due to the advances in deep learning, the FR technology shows high performance for most of the benchmark datasets. 
However, when the FR algorithm is applied to a real-world scenario, the performance has been known to be still unsatisfactory. 
This is mainly attributed to the mismatch between training and testing sets. 
Among such mismatches, face misalignment between training and testing faces is one of the factors that hinder successful FR.
To address this limitation, we propose a face shape-guided deep feature alignment framework for FR robust to the face misalignment. 
Based on a face shape prior (\textit{e.g.}, face keypoints), we train the proposed deep network by introducing alignment processes, \textit{i.e.}, pixel and feature alignments, between well-aligned and misaligned face images. Through the pixel alignment process that decodes the aggregated feature extracted from a face image and face shape prior, 
we add the auxiliary task to reconstruct the well-aligned face image. 
Since the aggregated features are linked to the face feature extraction network as a guide via the feature alignment process, we train the robust face feature to the face misalignment. 
Even if the face shape estimation is required in the training stage, the additional face alignment process, which is usually incorporated in the conventional FR pipeline, is not necessarily needed in the testing phase. 
Through the comparative experiments, we validate the effectiveness of the proposed method for the face misalignment with the FR datasets.  
\end{abstract}

\begin{IEEEkeywords}
Face recognition, face alignment, multi-task learning, face alignment learning, face shape prior.
\end{IEEEkeywords}}

\maketitle

\IEEEdisplaynontitleabstractindextext

%
\IEEEpeerreviewmaketitle

\IEEEraisesectionheading{\section{Introduction}\label{sec:introduction}}

%
%
%
%
\IEEEPARstart{T}{hanks} to the success of deep image classification and the availability of large-scale face image datasets, a deep face recognition (FR) has been actively studied. 
Recently, the FR performance has been considerably improved for many benchmark datasets. 
In particular, the performance for the LFW dataset~\cite{huang2008labeled} has shown about 99\% accuracy in face verification~\cite{deng2019arcface, wang2018cosface, liu2017sphereface}. 
However, when these FR algorithms are applied to the wild environment, the performance has been known to be highly degraded~\cite{masi2018learning, zhang2018improving, qian2019unsupervised}.
This is because benchmark datasets are usually collected from celebrities' face images with high quality, while face images in the wild environment suffer from the degradation of image quality (\textit{e.g.}, low-resolution, pose, and illumination variations). 
It leads to the mismatch of data distributions between the training and testing sets~\cite{kim2015face, an2019apa, zhou2018gridface, wong2010dynamic, choi2011comparative}. 

Among the mismatches, a face misalignment problem is one of the factors that hinder successful FR, which has been discussed in recent studies~\cite{an2019apa, zhou2018gridface, zhong2017toward, wu2017recursive,  wei2020balanced}.
For example, the face misalignment problem occurs with the testing face images not elaborately aligned or differently aligned with the training face images.
In order to deal with the face misalignment problem, most deep FR algorithms require the face alignment process based on the pre-defined canonical face location. 
For example, the recent deep FR algorithms~\cite{deng2019arcface, wang2018cosface, liu2017sphereface, wen2016discriminative} align face images by using fiducial points obtained from the multi-task convolutional neural network (MTCNN)~\cite{zhang2016joint} face detector.
Furthermore, deep face keypoint estimation algorithms~\cite{bulat2017far, bulat2018super, zhu2019robust} have been proposed for the elaborate face alignment.
However, the face keypoint estimation algorithm requires additional computational costs~\cite{lee2019lightweight}. 
In addition, the face detector or additional face keypoint estimation algorithms have inherent estimation errors, which cause the face misalignment problem.
More recently, to alleviate the issues related to the face alignment, there have been researches to learn a face alignment as well as a FR in an end-to-end manner (so-called \textit{face alignment learning})~\cite{an2019apa, zhou2018gridface,  zhong2017toward, wu2017recursive, wei2020balanced}.
By learning the face alignment with the FR simultaneously, these works have been validated to be effective under the face misalignment since a face image is automatically aligned to the proper alignment type in testing.
Motivated by the face alignment learning, in this paper, we improve the performance to be more robust to the face misalignment by learning the features being aware of face shape as well as a face image.

In this paper, we propose a face shape-guided deep feature alignment framework to address the face misalignment problem. 
Through both face images and the corresponding face shape priors (\textit{i.e.}, face keypoints), the face shape-guided feature is learned through pixel and feature alignment processes.
In detail, the aggregated feature from face images and face keypoints is utilized as guidance to align two features: feature considering only face image and feature considering face shape prior as well as face image (\textit{i.e.}, feature alignment). 
Then, the proposed deep network is collaboratively trained based on three tasks (\textit{i.e.},  face classification, pixel and feature alignments).
This shape-guided feature enables robust FR in test time without face alignment including keypoint estimation.
The contributions of the paper can be summarized as follows: 
\begin{itemize}[align=left,leftmargin=*,topsep=\parskip]
\setlength\itemsep{0.01em}
\item In training, by decoding the aggregated feature based on a face image and face shape prior, the proposed method learns the features for the well-aligned face image (\textit{i.e.}, pixel alignment). 
Through the feature alignment, the face feature extraction network as an input of only face image can learn the face shape-guided feature. 
\item In testing, a face feature vector invariant to face alignment is extracted only from a face image based on the trained network. 
Since our method does not require the explicit face alignment process, we can efficiently compute the robust feature to the face misalignment. 
\end{itemize}
\noindent Through the experiments, we verify the effectiveness of the proposed method conditioned under the face misalignment with face benchmark datasets.  

In Section~\ref{sec:related}, we briefly discuss the previous works. Section~\ref{sec:proposed} describes the proposed method. Then, the details of the experiments are presented in Section~\ref{sec:exp}. Finally, we conclude the paper in Section~\ref{sec:con}. 

\section{Related Work}\label{sec:related}

\subsection{Deep Face Recognition}\label{sec:relateddfr}

Recently, there have been many deep FR algorithms to effectively learn discriminative features using a convolutional neural network (CNN) since the DeepFace~\cite{taigman2014deepface}. 
The FaceNet~\cite{schroff2015facenet} was proposed to learn a Euclidean space embedding by introducing a triplet loss with 200 million face images. 
The authors in~\cite{wen2016discriminative} enhanced the discriminative power of the deeply learned features based on the center loss that penalizes the distances between the deep features and their corresponding class centers~\cite{wen2016discriminative}.
In order to learn angularly discriminative features, the SphereFace~\cite{liu2017sphereface} was proposed by introducing the angular SoftMax function (\textit{i.e.}, A-SoftMax) which imposes discriminative constraints on a hypersphere manifold~\cite{liu2017sphereface}. 
Also, the CosFace~\cite{wang2018cosface} with a large margin cosine loss was proposed to maximize further the decision margin in the angular space~\cite{wang2018cosface}. 
In~\cite{deng2019arcface}, the additive angular margin loss so-called ArcFace~\cite{deng2019arcface} was designed to obtain highly discriminative features for FR as well. 
More recently, to further improve FR performance, loss functions~\cite{duan2019uniformface,wei2020minimum,huang2019deep} and sample distribution-aware learning strategies~\cite{cao2020domain,huang2020curricularface} to enhance feature discrimination are being discussed.
Basically, all of these deep FR algorithms require the face alignment process to align a face image to a canonical view based on the facial keypoints.

\subsection{Facial Keypoint Estimation}\label{sec:relatedkeypoint}
As discussed earlier, the face alignment process based on the facial keypoint estimation is essential to learning the deep FR network, which has been actively studied. 
The MTCNN~\cite{zhang2016joint} has been proposed to jointly learn face keypoints as well as a face 
bounding box, where five facial keypoints (left/right eyes, nose, left/right mouse tips) are detected with face detection. 
Thanks to efficient computations, the MTCNN has been widely adopted for face image-based applications. 
To understand a face image more accurately, the face alignment network (FAN)~\cite{bulat2017far} was proposed for estimating $68$ facial keypoints based on the stacked hourglass 
network~\cite{newell2016stacked} that is widely used for human joint estimation. 
Furthermore, the FAN algorithm was extended to the 3D FAN algorithm for estimating 3D facial keypoints~\cite{bulat2017far}. 
Recently, the Super-FAN~\cite{bulat2018super} was proposed for estimating keypoints in the low-resolution face image, and the improved facial keypoint detector~\cite{zhu2019robust} robust to occlusion was proposed.
Despite the improved keypoint estimation performance, the keypoint estimation network is independently trained with the FR network, and it requires additional computational complexity~\cite{lee2019lightweight}. 
In addition, the facial keypoint estimation error caused by a low-quality face image promotes the vulnerability of the deep FR network to a face misalignment. 

\subsection{Face Alignment Learning}\label{sec:relatedfal}
To deal with the aforementioned limitations, recent studies were conducted to simultaneously train a face localization network for face alignment and a face feature extraction network for FR, which is called Face Alignment Learning.
Zhong \textit{et al.} proposed the end-to-end learning framework~\cite{zhong2017toward} for estimating the face image's transformation and FR by the Spatial Transformer Network (STN)~\cite{jaderberg2015spatial}.  
Here, STN's localization network predicted the 2D transform parameters of the face image to transform the face image into a canonical view.
However, the estimated transformation can capture only coarse geometric information as a holistic parametric model~\cite{zhou2018gridface}.
To deal with the accurate transformation (\textit{i.e.}, non-rigid transformation), the recursive spatial transformer (ReST)~\cite{wu2017recursive} by progressively aligning face images was proposed.
Despite the effective approach, the progressive aligning pipeline causes the degradation of the face image according to the repetitive rectifications for an image and features. 
Besides, GridFace~\cite{zhou2018gridface} was proposed to reduce geometric facial variations and improve the recognition performance by rectifying the face by local grid-level homography transformations~\cite{zhou2018gridface}. 
Recently, an adaptive pose alignment (APA)~\cite{an2019apa} has been proposed for aligning each face of training or test set to optimal alignment templates according to the facial pose instead of a predefined template~\cite{an2019apa}.
In~\cite{wei2020balanced}, Wei \textit{et al.} proposed the adaptive alignment of the face feature map based on the warp grid obtained from the localization network trained with face keypoints.
In this paper, we propose the end-to-end face feature learning framework guided by the face shape as the face alignment learning instead of a direct transformation of a face image or feature maps. 

\begin{figure*}[!t]
\centering
\includegraphics[width=0.9\linewidth]{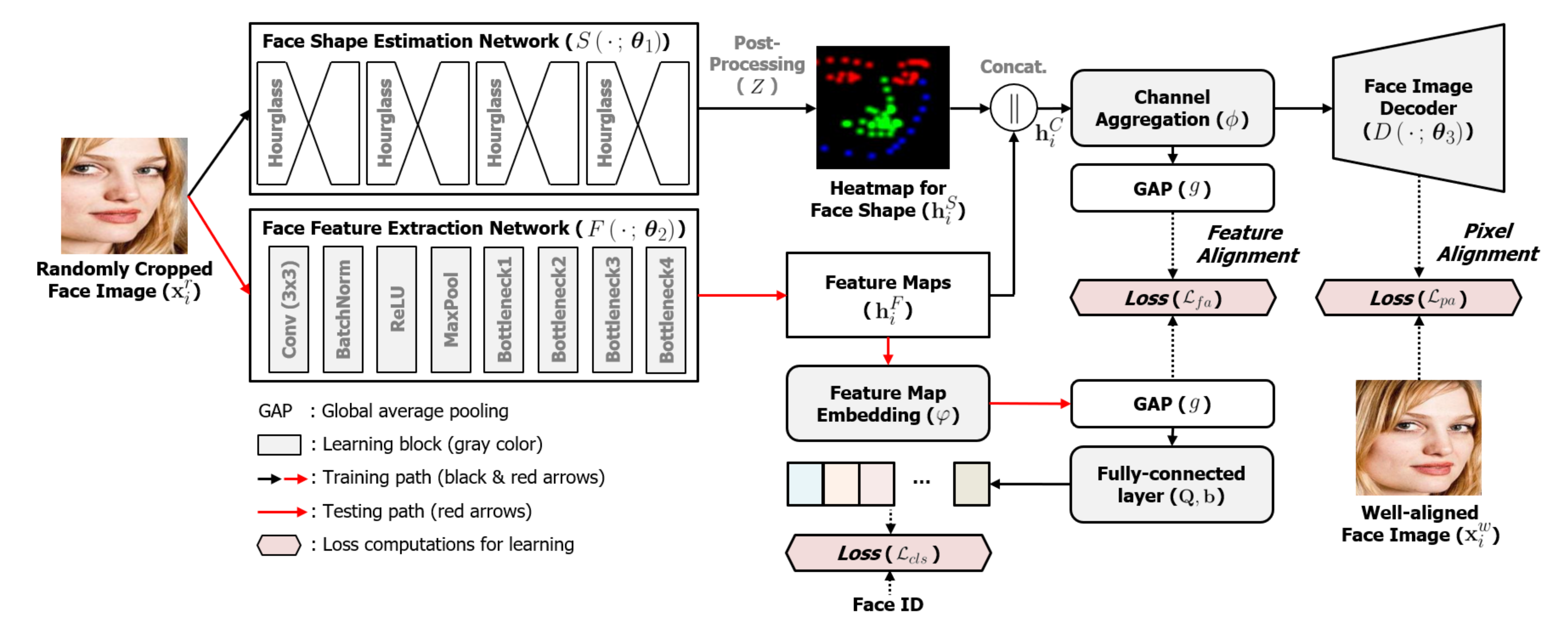}
\begin{center}
\end{center}
\vspace{-1.1cm}
   \caption{Overview of the proposed face shape-guided deep feature alignment framework. The proposed method is mainly comprised of the face shape estimation network, face feature extraction network, and face image decoder. Based on three networks, two alignment processes (\textit{i.e.}, pixel alignment and feature alignment) are introduced. 
   Then, through the pixel alignment and feature alignment processes, the face shape-guided feature is trained by jointly classifying the face and reconstructing the well-aligned face image. Note that only the face feature extraction network is used for the testing phase. 
   Best viewed in color.}
\label{fig_proposed}
\vspace{-0.5cm}
\end{figure*}

\section{Proposed Method}\label{sec:proposed}
In this paper, we propose a face shape-guided deep feature alignment framework robust to face misalignment. 
By using both a randomly cropped face image ($\mathbf{x}^{r}_{i}$) from the $i$-th training face image ($\mathbf{x}_{i}$) and the corresponding face shape prior, the face shape-guided feature is trained based on the well-aligned face images ($\mathbf{x}^{w}_{i}$). 
Note that the well-aligned face images denote face images cropped by considering face keypoints. And, the randomly cropped face images are obtained by over-sampling strategy~\cite{cao2018vggface2}. Please please refer to Preprocessing in Section~\ref{sec:expset} for details.
When testing, our face shape-guided feature vector, which is invariant to the face alignment, enables us to achieve robust FR to face misalignment without the additional face alignment process. 
In the following subsections, we describe more details of the proposed method. 

\subsection{Architecture}\label{sec:proposedarch}
As shown in Fig.~\ref{fig_proposed}, the proposed method mainly consists of three parts for the purpose of face feature extraction robust to face misalignment: 1) Face shape estimation network ($S$), 2) Face feature extraction network ($F$), and 3) Face image decoder ($D$), which are connected by two separated neural layers (\textit{i.e.}, feature map embedding ($\varphi$) and channel aggregation ($\phi$)). 
The face shape estimation network infers the coordinates of the facial keypoints as a face shape prior, resulting in the face shape's heatmaps. 
And, the face feature extraction network outputs face appearance features based on a backbone network. 
After aggregating outputs from the face shape estimation and feature extraction networks, the well-aligned face image is reconstructed by the face image decoder from the randomly cropped face image (\textit{i.e.}, pixel alignment). 
Through the pixel alignment, we can model the characteristic of the well-aligned face image.
The face shape-aggregated feature is simultaneously connected to the face feature extraction network through the feature alignment.
It helps our face feature extraction network to learn the face shape-guided feature. 
%
In other words, the feature alignment is helpful for extracting the face shape-guided feature without the help of the face shape estimation network in testing.
By learning the characteristic of the well-aligned face image in an end-to-end manner, we can extract face features robust to face misalignment in the testing phase without an explicit face alignment process. 

\subsubsection{Face Shape Estimation Network}
In order to estimate the face shape prior, we use the 2D face alignment network (FAN)~\cite{bulat2017far}, which consists of four consecutive hourglass modules~\cite{newell2016stacked}. 
In this paper, the FAN architecture and the corresponding pre-trained model are utilized for the face shape estimation without any modification, where the pre-trained parameters are frozen in training to obtain the face shape prior stably. 
For the input face image ($\mathbf{x}^{r}_{i}$, \textit{i.e.}, randomly cropped face image), the network parameterized by $\boldsymbol{\theta}_{1}$ infers $68$-channel heatmaps for the corresponding $68$ facial keypoints. 
To take advantage of the heatmaps effectively, we perform the following post-processing method: Gaussian blurring, resizing, and channel conversion. 
First, the $68$-channel heatmaps are blurred by a Gaussian kernel with $\sigma$ for the location of the heatmap’s peak point (\textit{i.e.}, the facial keypoint location). This is to emphasize the importance of the surrounding areas around the keypoints as suggested in~\cite{Khorramshahi_2019_ICCV}. 
Then, we resize the heatmaps to $56\times56$ to align with the feature maps extracted from the face feature extraction network. 
Note that the 68-channel heatmaps for the estimated keypoints are converted to 3-channel heatmaps for efficient memory consumption. 
In other words, the converted heatmap is made to have three image channels, where the channel corresponds to each part: 1) R-channel: heatmap for eyes and eyebrows, 2) G-channel: heatmap for the nose and mouth, and 3) B-channel: heatmap for the face boundary as shown in Fig.~\mbox{\ref{fig_proposed}}.
In summary, the extracted heatmap ($\mathbf{h}_{i}^{S}$) is represented as follows:
\begin{equation}
\mathbf{h}^{S}_{i}=Z\left[S\left(\mathbf{x}^{r}_{i};\,\boldsymbol{\theta}_{1}\right)\right],
\label{eq_shape}
\end{equation}
where $S$ denotes the operation of the FAN’s feedforward computation. And, $Z$ means the post-processing function for the Gaussian blurring, resizing, and channel conversion. 

\subsubsection{Face Feature Extraction Network}
To extract face appearance features, we design the face feature extraction network ($F$) by modifying the ResNet50~\cite{he2016deep}. 
Like the original ResNet50, the $F$ is comprised of 2D convolution, batch normalization, ReLU activation, max-pooling layers, and four Bottleneck layers~\cite{he2016deep} in order. 
Here, the number of channels in each layer and the stride in the Bottleneck layers are modified to align feature maps with the heatmaps for the face shape prior. 
Specifically, the number of channels is all halved, \textit{i.e.}, $512$ channels are changed to $256$ channels in the fourth Bottleneck layer. 
And, the stride for the bottleneck layer is set to $1$ instead of $2$. 
For the $\mathbf{x}^{r}_{i}$, the face appearance feature maps ($\mathbf{h}^{F}_{i}$) are obtained by the following equation: 
\begin{equation}
\mathbf{h}^{F}_{i}=F\left(\mathbf{x}^{r}_{i};\,\boldsymbol{\theta}_{2}\right),
\label{eq_feat}
\end{equation}
where $\boldsymbol{\theta}_{2}$ is the parameters of the face feature extraction network.

\begin{figure}[!t]
  \centering
 \includegraphics[width=0.8\linewidth]{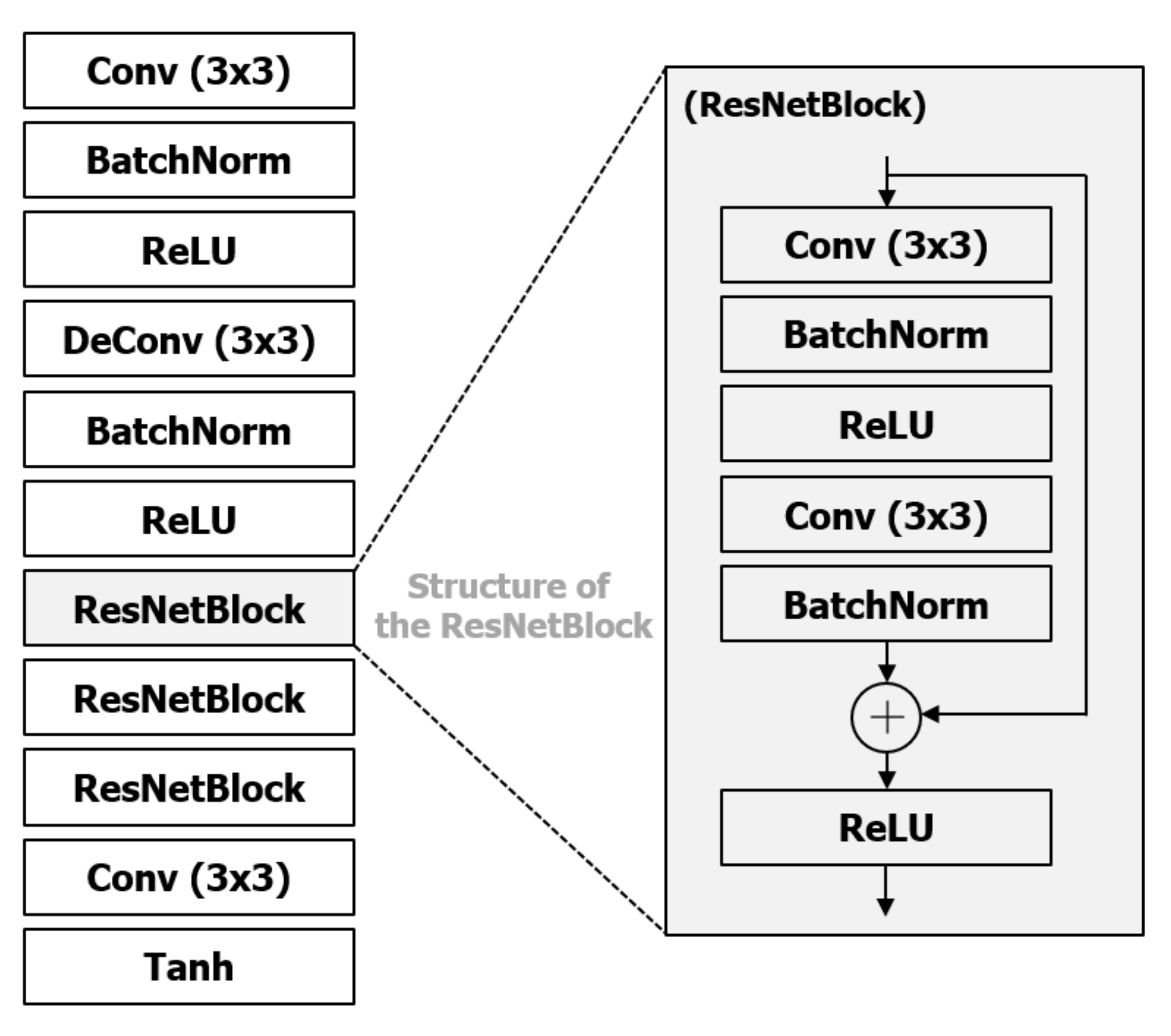}
 \vspace{-0.3cm}
 \caption{The structure of the face image decoder ($D$), where the structure of the ResNetBlock (\textit{i.e.}, Basic block~\cite{he2016deep} in the $D$ is presented in the gray box.). Note that the `Conv' and 'DeConv' denote convolution and deconvolution operations, respectively.}
 \vspace{-0.5cm}
\label{fig_dec}
\end{figure}
\subsubsection{Face Image Decoder}
The face image decoder is incorporated to learn the characteristic of the well-aligned face image $\mathbf{x}^{w}_{i}$ by using features from $\mathbf{h}^{S}_{i}$ and $\mathbf{h}^{F}_{i}$.
Prior to the decoding, both the $\mathbf{h}^{S}_{i}$ and $\mathbf{h}^{F}_{i}$ are concatenated to form a stacked feature map ($\mathbf{h}^{C}_{i}$) as: 
\begin{equation}
\mathbf{h}^{C}_{i}=\left[\mathbf{h}^{S}_{i}\parallel\mathbf{h}^{F}_{i}\right],
\label{eq_concat}
\end{equation}
where ``$\parallel$'' symbol means the concatenation operation in a channel direction. 
Then, the stacked feature $\mathbf{h}^{C}_{i}$ is effectively aggregated to make a face shape-guided feature through the Channel Aggregation. 
The channel aggregation layer is simply comprised of one $1\times1$ convolution layer and one batch normalization layer, which fuses face appearance features and face shape prior in a channel-wise manner.
For simplicity, we denote the channel aggregation operation as $\phi$. 
The input of the face image decoder can be represented as $\phi(\mathbf{h}^{C}_{i})$. 
The decoded result is obtained by the following equation: 
\begin{equation}
\tilde{\mathbf{x}}_{i}^{w}=D\left(\phi\left(\mathbf{h}^{C}_{i}\right);\,\boldsymbol{\theta}_3\right),
\label{eq_recon}
\end{equation}
where the face image decoder ($D$) is parameterized by $\boldsymbol{\theta}_3$. 
The decoder network firstly reduces the number of concatenated feature maps to $64$ by a $3\times3$ convolution layer. 
The next $3\times3$ deconvolution layer is used for upsampling the feature map to double the resolution.
Then, three residual blocks are used to decode features. 
Finally, a $3\times3$ convolution layer reconstructs the well-aligned face image. 
The detailed architecture of the face image decoder is shown in Fig.~\ref{fig_dec}. 
Note that the process to reconstruct the well-aligned face image from the stacked features of the randomly cropped face image is called to \textit{Pixel Alignment}. 

In addition to the procedure mentioned above, an additional learning path should be devised to extract face features that are robust to face misalignment in the testing. 
Motivated by the concept of the feature alignment in the recent researches~\cite{suh2018part, bozorgtabar2019syndemo}, we design the learning path for the \textit{Feature Alignment}. 
In other words, the feature alignment in our network enables us to train the face feature extraction network guided by the channel aggregated feature used for the face image decoder. 
Since the channel aggregated feature includes information related to the face appearance and the face shape, the additional learner $\varphi$ learns the function to map the face appearance feature $\mathbf{h}^{F}_{i}$ extracted from the face feature extraction network $F$ to the guidance ($\phi(\mathbf{h}^{C}_{i})$). 
Note that we transform both feature maps $\phi(\mathbf{h}^{C}_{i})$ and $\varphi(\mathbf{h}^{F}_{i})$ to feature vectors by the global average pooling (GAP) for efficient computations~\cite{lin2013network}.
In Table~\ref{tab_arch}, the details of the architecture to be trained in our framework are summarized.

\begin{table*}
\begin{center}
\caption{Summary of the proposed deep network's architecture. $[\cdot]$ represents the operations of the ResNet blocks (Bottleneck~\cite{he2016deep} and Basic~\cite{he2016deep}). DeConv layer means the deconvolution operation to enlarge the resolution of a feature map. And, $8631$ in the fully connected layer stands for the number of classes in the training dataset.}
\label{tab_arch}
\vspace{-0.3cm}
\begin{tabular}{|c||c|c|c|}
\hline
\textbf{Network}               & \textbf{Layer Name}                                                      & \textbf{Output Size}  & \textbf{Operation} \\\Xhline{3\arrayrulewidth}
\multirow{6}{*}[-9.5ex]{\begin{tabular}[c]{@{}c@{}}Face Feature \\ Extraction Network\\($F$)\end{tabular}} &
  Conv1 &
  $112\times112$ &
  \begin{tabular}[c]{@{}c@{}}$7\times7$, $64$, $\mathrm{stride}$\,$2$\\ (BatchNorm, ReLU)\end{tabular}
  \\ \cline{2-4} 
 &
  \multirow{2}{*}[-3.0ex]{\begin{tabular}[c]{@{}c@{}}Conv2\_x\\ (Bottleneck)\end{tabular}} &
  \multirow{2}{*}[-3.0ex]{$56\times56$} &$3\times3$ Max Pooling, $\mathrm{stride}$\,$2$
   \\  \cline{4-4} 
                      &                                                                 &              &           $\begin{bmatrix}
1\times1,\,32\,,\mathrm{stride}\,1\\ 
3\times3,\,32\,,\mathrm{stride}\,3\\
1\times1,\,128\,,\mathrm{stride}\,1
\end{bmatrix}\times3$\\ \cline{2-4} 
                      & \begin{tabular}[c]{@{}c@{}}Conv3\_x\\ (Bottleneck)\end{tabular} & $56\times56$ &    $\begin{bmatrix}
1\times1,\,64\,,\mathrm{stride}\,1\\ 
3\times3,\,64\,,\mathrm{stride}\,3\\
1\times1,\,256\,,\mathrm{stride}\,1
\end{bmatrix}\times4$       \\ \cline{2-4} 
                      & \begin{tabular}[c]{@{}c@{}}Conv4\_x\\ (Bottleneck)\end{tabular} & $56\times56$ &     $\begin{bmatrix}
1\times1,\,128\,,\mathrm{stride}\,1\\ 
3\times3,\,128\,,\mathrm{stride}\,3\\
1\times1,\,512\,,\mathrm{stride}\,1
\end{bmatrix}\times6$      \\ \cline{2-4} 
                      & \begin{tabular}[c]{@{}c@{}}Conv5\_x\\ (Bottleneck)\end{tabular} & $56\times56$ &   $\begin{bmatrix}
1\times1,\,256\,,\mathrm{stride}\,1\\ 
3\times3,\,256\,,\mathrm{stride}\,3\\
1\times1,\,1024\,,\mathrm{stride}\,1
\end{bmatrix}\times3$        \\ \hline
\multirow{6}{*}[-7.0ex]{\begin{tabular}[c]{@{}c@{}}Face Image \\ Decoder\\($D$)\end{tabular}} &
  Conv1 & $56\times56$
   & \begin{tabular}[c]{@{}c@{}}$3\times3$, $64$, $\mathrm{stride}$\,$1$\\ (BatchNorm, ReLU)\end{tabular}
   \\ \cline{2-4} 
                      & DeConv                                                          &  $112\times112$            &   \begin{tabular}[c]{@{}c@{}}$3\times3$, $64$, $\mathrm{stride}$\,$2$\\ (BatchNorm, ReLU)\end{tabular}        \\ \cline{2-4} 
                      & \begin{tabular}[c]{@{}c@{}}Conv2\_x\\ (Basic)\end{tabular}                                                        &  $112\times112$            &   $\begin{bmatrix}
3\times3,\,64\,,\mathrm{stride}\,1\\ 
3\times3,\,64\,,\mathrm{stride}\,1
\end{bmatrix}\times2$        \\ \cline{2-4} 
                      & \begin{tabular}[c]{@{}c@{}}Conv3\_x\\ (Basic)\end{tabular}                                                        &  $112\times112$             &   $\begin{bmatrix}
3\times3,\,64\,,\mathrm{stride}\,1\\ 
3\times3,\,64\,,\mathrm{stride}\,1
\end{bmatrix}\times2$        \\ \cline{2-4} 
                      & \begin{tabular}[c]{@{}c@{}}Conv4\_x\\ (Basic)\end{tabular}                                                        &  $112\times112$             &   $\begin{bmatrix}
3\times3,\,64\,,\mathrm{stride}\,1\\ 
3\times3,\,64\,,\mathrm{stride}\,1
\end{bmatrix}\times2$        \\ \cline{2-4} 
                      & Conv5                                                           &    $112\times112$          &  \begin{tabular}[c]{@{}c@{}}$3\times3$, $3$, $\mathrm{stride}$\,$1$\\ (Tanh)\end{tabular}         \\ \hline

Feature Map Embedding ($\varphi$)     & Conv &              $56\times56$&   \begin{tabular}[c]{@{}c@{}}$1\times1$, $512$, $\mathrm{stride}$\,$1$\\ (BatchNorm, ReLU)\end{tabular}        \\ \hline

Channel Aggregation ($\phi$)   & Conv                                                            &              $56\times56$&   \begin{tabular}[c]{@{}c@{}}$1\times1$, $512$, $\mathrm{stride}$\,$1$\\ (BatchNorm)\end{tabular}        \\ \hline
Fully Connected Layer ($\mathbf{Q}$, $\mathbf{b}$) & -                                                               &              $1\times1$&  \begin{tabular}[c]{@{}c@{}}Global Average Pooling, 8631\\ (SoftMax)\end{tabular}     \\ \hline
\end{tabular}
\end{center}
\end{table*}

\begin{algorithm}
\DontPrintSemicolon
\KwIn{Randomly cropped and well-aligned face image pairs and corresponding ID labels, Parameters $\boldsymbol{\theta}_{1}$ for Face Shape Estimation Network, Number of epochs $n_{E}$, Number of batches $n_B$, Learning rate $\delta$, Momentum $\tau$}
\KwOut{$\boldsymbol{\Theta}=\left \{\boldsymbol{\theta}_{2}, \boldsymbol{\theta}_{3}, \phi, \varphi, \mathbf{Q}, \mathbf{b}\right \}$}
\SetKwBlock{Begin}{function}{end function}
\For{$t=1,\cdots,n_{E}$}
{
	\For{$b=1,\cdots,n_{B}$}
	{
		\# \textit{Feed-forward Operation}\;
		Forward propagating $S$ by $\boldsymbol{\theta}_{1}$ and post -processing $Z$ to obtain face shape priors; \;
		Forward propagating $F$ by $\boldsymbol{\theta}_{2}$ to obtain the face feature maps; \;
		Concatenating the face shape prior and face feature maps, then forward propagating $\phi$;\;
		Forward propagating $\varphi, \mathbf{Q}, \mathbf{b}$ to obtain the estimated ID labels; \;
		Forward propagating $D$ by $\boldsymbol{\theta}_{3}$ to obtain the reconstructed face images;\;\;

		\# \textit{Loss Computation}\;
		Compute $\mathcal{L}_{cls}$ by Eq.~(\ref{eq_cls});\;
		Compute $\mathcal{L}_{pa}$ by Eq.~(\ref{eq_pa});\;
		Compute $\mathcal{L}_{fa}$ by Eq.~(\ref{eq_fa});\;
		Weighted sum ($\mathcal{L}$) of loss functions by Eq.~(\ref{eq_loss});\;\;
		
		\# \textit{Parameter Update}\;
		\eIf{$b=1$}
		{
		    $\mathbf{v}_{b}^{t}\leftarrow\nabla_{\boldsymbol{\Theta}}\mathcal{L}$;\;
		}{
			$\mathbf{v}_{b}^{t}\leftarrow\tau \mathbf{v}_{b-1}^{t}+\nabla_{\boldsymbol{\Theta}}\mathcal{L}$;\;
	    }
		Update: $\boldsymbol{\Theta}_{b+1}^{t}\leftarrow\boldsymbol{\Theta}_{b}^{t}-\delta^{t}\mathbf{v}_{b}^{t}$;

	}
	$\boldsymbol{\Theta}_{1}^{t+1}\leftarrow\boldsymbol{\Theta}_{n_{B}}^{t}$}

%
%
%
\caption{Pseudo code for training the proposed method. All training samples are divided into $n_{B}$ batches and used for training.}\label{alg_proposed}
\end{algorithm}
\subsection{Training}\label{sec:proposedtrn}
In order to train the proposed network, three-loss functions are introduced. 
First, a cross-entropy loss is used to classify a face image into one of the classes for the embedded feature $g\left(\varphi\left(\mathbf{h}^{F}_{i}\right)\right)$ after the feature extraction by $F$ and the GAP, which is defined by  
\begin{equation}
\mathcal{L}_{cls}=-\frac{1}{nc}\sum_{\forall i}\sum_{\forall c}y_{i}^{c}\log\tilde{y}_{i}^{c},
\label{eq_cls}
\end{equation}
where $y_{i}^{c}\in\left \{0,1\right \}$ is the $c$-th element of the one-hot vector corresponding to the ground truth class label of the $i$-th sample, and $\tilde{y}_{i}^{c}$ is the $c$-th element of the estimated label by a SoftMax function, \textit{i.e.}, $\mathrm{SoftMax}\left(\mathbf{Q}^{\top}g\left(\varphi\left(\mathbf{h}^{F}_{i}\right)\right)+\mathbf{b}\right)$. 
The $\mathbf{Q}$ and $\mathbf{b}$ are the weight matrix and bias vector for the fully-connected layer. 
The $n$ and $c$ are the number of samples used for training in an epoch and the number of classes, respectively. 
Then, to learn the face image decoder, the L1 loss between the well-aligned face image and the decoded face image, \textit{i.e.}, loss for the pixel alignment, is defined by 
\begin{equation}
\mathcal{L}_{pa} = \frac{1}{n}\sum_{\forall i}\left \| \mathbf{x}^{w}_{i}-\tilde{\mathbf{x}}_{i}^{w}\right \|_{1}, \,\mathrm{where}\,\, \tilde{\mathbf{x}}_{i}^{w}=D\left(\phi\left(\mathbf{h}^{C}_{i}\right);\,\boldsymbol{\theta}_{3}\right).
\label{eq_pa}
\end{equation} 
Finally, the loss for the feature alignment to learn the face shape-guided feature is defined by the following equation: 
\begin{equation}
\mathcal{L}_{fa} = \frac{1}{n}\sum_{\forall i}\left \| g\left(\phi\left(\mathbf{h}^{C}_{i}\right)\right)-g\left(\varphi\left(\mathbf{h}^{F}_{i}\right)\right) \right \|_{2}^{2}.
\label{eq_fa}
\end{equation}
From Eq.~(\ref{eq_fa}), the feature considering both the face image and face shape prior, $g\left(\phi\left(\mathbf{h}^{C}_{i}\right)\right)$, is distilled to the feature from the face feature extraction network, $g\left(\varphi\left(\mathbf{h}^{F}_{i}\right)\right)$. 
By using three loss functions, the total loss ($\mathcal{L}$) for training the proposed deep network is defined as: 
\begin{equation}
\mathcal{L} = \alpha\mathcal{L}_{cls} + \beta\mathcal{L}_{pa} + \gamma\mathcal{L}_{fa}, 
\label{eq_loss}
\end{equation}
where $\alpha$, $\beta$, and $\gamma$ denote the hyper-parameters that control the balance for the total loss function.
We optimize the following objective function to obtain the learning parameters (\textit{i.e.}, $\boldsymbol{\Theta} =\left \{ \boldsymbol{\theta}_{2},\boldsymbol{\theta}_{3},\phi, \varphi, \mathbf{Q}, \mathbf{b} \right \}$) by mini-batch gradient descent:
\begin{equation}
\boldsymbol{\Theta}^{*}=\textup{arg}\,\underset{\boldsymbol{\Theta}}{\textup{min}}\,\,\mathcal{L}.
\label{eq_opt}
\end{equation}
Based on three-loss functions, we train the proposed deep network with three steps: 1) feed-forward operation, 2) loss computation, and 3) parameter update by mini-batch gradient descent. 
The training method is summarized in Algorithm~\ref{alg_proposed}.

\begin{figure}[!t]
  \centering
 \includegraphics[width=1.0\linewidth]{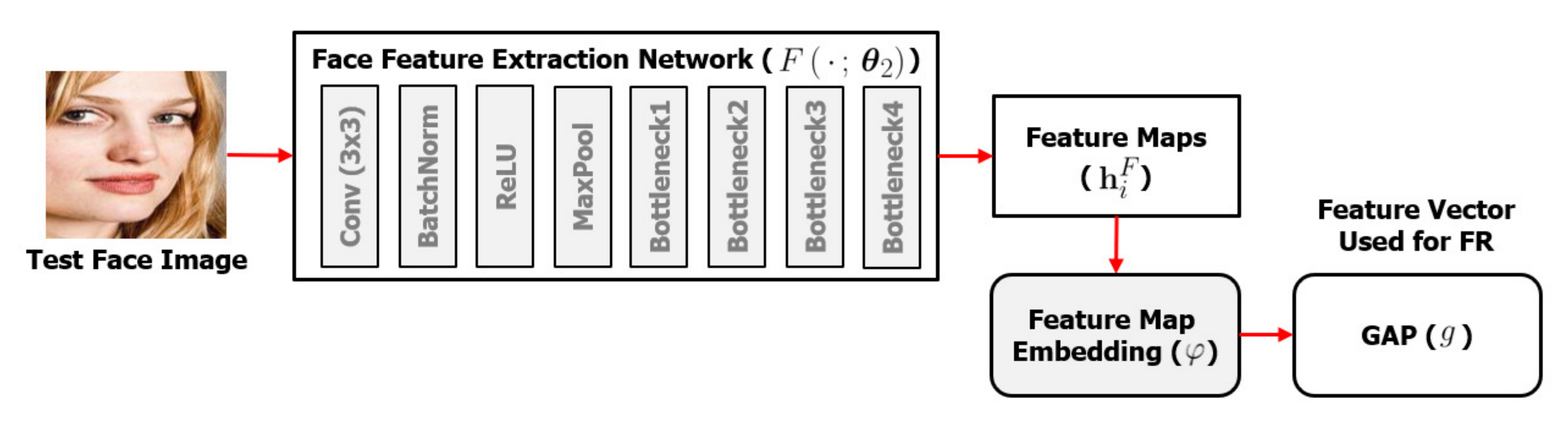}
 \vspace{-0.3cm}
 \caption{Inference step for the proposed method.}
\label{fig_inf}
\end{figure}
\subsection{Inference}\label{sec:proposedinf}
As can be seen in Fig.~\ref{fig_inf}, given the learned face feature extraction network $F$, an input face image is processed to extract a feature vector (\textit{i.e.}, $g(\varphi(\mathbf{h}^{F}))$) robust to the face misalignment regardless of the networks $S$ and $D$. 
The extracted feature vector is used for the FR. 
Note that even if the proposed face shape-guided deep feature alignment framework requires a face shape prior extracted by $S$, the face shape prior is no longer required in the testing step. 
Thus, it can reduce the computational complexity compared to the conventional FR pipeline equipped with an explicit face alignment process.  

\begin{table}[!t]
\caption{Summary of data preprocessing and hyper-parameters for training the proposed network. ($^{\dagger}$ All face images were normalized to have values ranging from -1 to 1, dividing 128.0 after subtracting 127.5 from original pixels values after the preprocessing. $^{\ast}$ The learning rate ($\delta$) was decayed by the factor of 0.1 for every 30 epochs.)}
\label{tab_set}
\centering
\vspace{-0.3cm}
\resizebox{\linewidth}{!}{%
\begin{tabular}{|c|c|c|c|}
\hline
\multicolumn{4}{|c|}{\textbf{Data preprocessing$^\dagger$}}             \\ \Xhline{3\arrayrulewidth} 
\multicolumn{4}{|l|}{\underline{Well-aligned face image ($\mathbf{x}^{w}$)}}             \\  
\multicolumn{4}{|l|}{- Crop original face image based on the bounding box}             \\  
\multicolumn{4}{|l|}{- Align face image based on the keypoints via similarity transform}             \\  
\multicolumn{4}{|l|}{- Resize the face image to 224$\times$224 pixels}             \\  
\multicolumn{4}{|l|}{}             \\ 
\multicolumn{4}{|l|}{\underline{Randomly cropped face image ($\mathbf{x}^{r}$)}}             \\  
\multicolumn{4}{|l|}{- Adjust bounding box coordinates with 10 pixels margin}             \\  
\multicolumn{4}{|l|}{- Resize the face image to 256$\times$256 pixels}             \\  
\multicolumn{4}{|l|}{- Randomly crop 224$\times$224 pixels within the 256$\times$256 pixels face image}             \\  \hline\hline
\multicolumn{4}{|c|}{\textbf{Hyper-parameters for training the proposed deep network}}             \\  
\Xhline{3\arrayrulewidth}
Learning rate$^{\ast}$ ($\delta$)              & 0.1            & Momentum ($\tau$)  &           0.9\\
Number of epochs ($n_{E}$)             & 100            & Number of batches ($n_{B}$)  &           64\\
Weight for $\mathcal{L}_{cls}$ ($\alpha$)              & 1.0            & Weight for $\mathcal{L}_{pa}$ ($\beta$)  &           1.0\\
Weight for $\mathcal{L}_{fa}$ ($\gamma$)              & 1.0            & Gaussian blurring ($\sigma$)  &           2.0\\\hline
\end{tabular}%
}
\end{table}
\begin{figure*}[!t]
\centering
\includegraphics[width=1.0\linewidth]{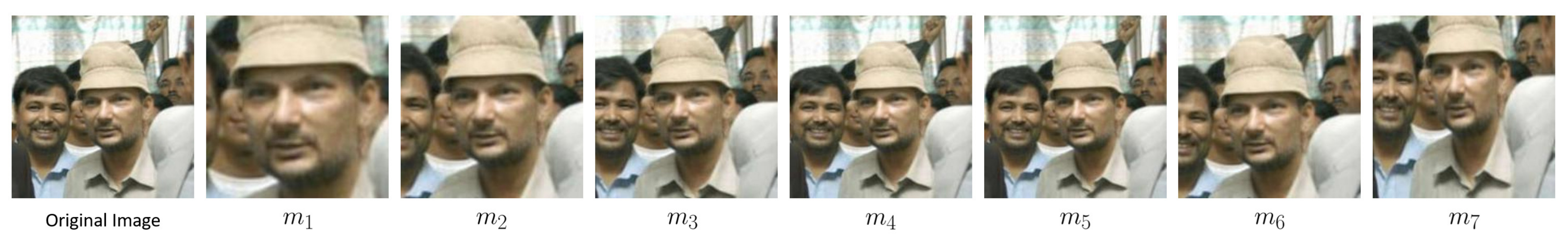}
\begin{center}
\end{center}
\vspace{-1.1cm}
   \caption{Example face images used for the experiments in order to investigate the effect of the face misalignment depending on the different margin parameters.}
\label{fig_align}
\vspace{-0.5cm}
\end{figure*}
\section{Experiments}\label{sec:exp}
\subsection{Experimental Settings}\label{sec:expset}
\textbf{Dataset.} For the experiments, the large-scale VGGFace2~\cite{cao2018vggface2} dataset was used for training the proposed method. 
The VGGFace2 dataset is comprised of $3.31$ million face images from $9,131$ identities which have large variations in pose, age, illumination, ethnicity, and profession~\cite{cao2018vggface2}. 
As suggested in~\cite{cao2018vggface2}, face images from $8,631$ persons were used for training the proposed deep network, and the remaining face images for $500$ disjoint persons were used for the validation. 
To evaluate the proposed method, we adopted four benchmark datasets: 1) LFW~\cite{huang2008labeled}, 2) Cross-Age LFW (CALFW)~\cite{zheng2017cross}, 3) Cross-Pose LFW (CPLFW)~\cite{zheng2018cross}, 4) YTF~\cite{wolf2011face}, and 5) MegaFace~\cite{kemelmacher2016megaface}. 
The LFW dataset contains $13,233$ face images from $5,749$ face identities and provides $6,000$ pairs for face verification test~\cite{huang2008labeled}. 
As the reorganized version of the LFW dataset, the CALFW dataset is comprised of $6,000$ pairs with large-age variations~\cite{zheng2017cross} for face verification tests as well.
The CPLFW dataset was constructed by searching and selecting 3,000 positive face pairs with pose differences to add pose variation to intra-class variance~\cite{zheng2018cross}.
And 3,000 negative pairs with the same gender and race were also selected to reduce the influence of attribute difference between positive and negative pairs~\cite{zheng2018cross}.
The YTF dataset includes $3,425$ videos of $1,595$ identities, where two video clips are verified whether they are matched or not~\cite{wolf2011face}. 
For a face verification with four datasets, we measured accuracy with 10-fold cross-validation.
For a face identification, we used MegaFace dataset which consists of $1$ million distractors with the Challenge $1$ of $100,000$ face images from $530$ celebrities (\textit{i.e.}, FaceScrub~\cite{ng2014data}). 
Note that the cleaned version of the MegaFace dataset was adopted in our experiment as introduced in~\cite{deng2019arcface}. 
For all experiments, the $512$-dimensional feature vectors for input face images were extracted, then normalized feature vectors to have a unit norm by a L2 normalization. 
The distance between face images in verification pairs (LFW, CALFW, and YTF), as well as the gallery-probe list (MegaFace), was computed based on a cosine distance. 

\textbf{Preprocessing.}
As discussed earlier, two types of face images were used for training. 
The well-aligned face image ($\mathbf{x}^{w}$) was obtained by cropping an original face image based on the bounding box coordinates provided in the VGGFace2 dataset and keypoints by resizing the face images to $224\times224$ pixels. 
In contrast, the randomly cropped face images ($\mathbf{x}^{r}$) were obtained by adjusting bounding box coordinates with $10$ pixels margin, then resized to $256\times256$ pixels. 
Finally, the face images were randomly cropped to have $224\times224$ pixels. 
Pixel intensities were normalized to have values ranging from $-1$ to $1$ by dividing $128.0$ after subtracting $127.5$ from original pixel values.

\begin{table*}[!t]
\caption{Face verification performance (Accuracy, \%) for the LFW dataset depending on the margin parameters, where ``Optimal Alignment'' means that the required optimal alignment is satisfied for each deep FR method.}
\label{tab_lfw2}
\centering
\vspace{-0.3cm}
\resizebox{\textwidth}{!}{%
\begin{tabularx}{\linewidth}{|Y||Y|Y|Y|Y|Y|Y|Y|Y|}
\hline
\multirow{2}{*}{} & \multicolumn{7}{c|}{Margin parameters to control face alignment} & \multirow{2}{*}{\begin{tabular}[c]{@{}c@{}}Optimal\\ Alignment\end{tabular}}\\\cline{2-8} 
                   & $m_1$                 & $m_2$                 & $m_3$                 & $m_4$                 & $m_5$ & $m_6$ & $m_7$ & \\ 
\Xhline{3\arrayrulewidth}
\multicolumn{1}{|c||}{SphereFace~\cite{liu2017sphereface}}   
 &\multicolumn{1}{c|}{86.82$\pm$ 1.34} &\multicolumn{1}{c|}{79.55$\pm$ 2.33} &\multicolumn{1}{c|}{75.15$\pm$ 1.96}     &\multicolumn{1}{c|}{72.85$\pm$ 2.35} &\multicolumn{1}{c|}{72.37$\pm$ 2.66}                 
&\multicolumn{1}{c|}{73.47$\pm$ 2.30}       &\multicolumn{1}{c|}{75.15$\pm$ 1.80}    & \multicolumn{1}{c|}{99.10$\pm$ 0.38}      \\ 

\multicolumn{1}{|c||}{CosFace~\cite{wang2018cosface}}       &\multicolumn{1}{c|}{97.25$\pm$ 1.10}     
&\multicolumn{1}{c|}{93.40$\pm$ 1.07} &\multicolumn{1}{c|}{85.55$\pm$ 1.55}                 &\multicolumn{1}{c|}{80.77$\pm$ 1.76}                 &\multicolumn{1}{c|}{80.20$\pm$ 1.77}       &\multicolumn{1}{c|}{82.95$\pm$ 1.11}         &\multicolumn{1}{c|}{82.75$\pm$ 2.07}& \multicolumn{1}{c|}{99.52$\pm$ 0.30}\\

\multicolumn{1}{|c||}{VGGFace2~\cite{cao2018vggface2}}      &\multicolumn{1}{c|}{98.98$\pm$ 0.55}     
&\multicolumn{1}{c|}{95.95$\pm$ 0.84} &\multicolumn{1}{c|}{84.63$\pm$ 1.86}                 &\multicolumn{1}{c|}{77.85$\pm$ 1.89}                 &\multicolumn{1}{c|}{77.87$\pm$ 1.91}       &\multicolumn{1}{c|}{83.50$\pm$ 1.83}         &\multicolumn{1}{c|}{86.75$\pm$ 1.26}& \multicolumn{1}{c|}{99.08$\pm$ 0.54}\\

\multicolumn{1}{|c||}{ArcFace~\cite{deng2019arcface}}       &\multicolumn{1}{c|}{94.72$\pm$ 0.70}     
&\multicolumn{1}{c|}{93.12$\pm$ 1.29} &\multicolumn{1}{c|}{84.38$\pm$ 1.44}                 &\multicolumn{1}{c|}{79.95$\pm$ 1.99}                 &\multicolumn{1}{c|}{79.03$\pm$ 2.08}       &\multicolumn{1}{c|}{83.07$\pm$ 1.80}         &\multicolumn{1}{c|}{76.87$\pm$ 2.28}& \multicolumn{1}{c|}{\textbf{99.72$\pm$ 0.19}}\\
\Xhline{1\arrayrulewidth}
\multicolumn{1}{|c||}{Proposed}       &\multicolumn{1}{c|}{\textbf{99.28$\pm$ 0.58}}     
&\multicolumn{1}{c|}{\textbf{99.30$\pm$ 0.44}} &\multicolumn{1}{c|}{\textbf{98.85$\pm$ 0.65}}                 &\multicolumn{1}{c|}{\textbf{97.47$\pm$ 1.06}}                 &\multicolumn{1}{c|}{\textbf{97.12$\pm$ 1.10}}       &\multicolumn{1}{c|}{\textbf{98.80$\pm$ 0.79}}         &\multicolumn{1}{c|}{\textbf{98.67$\pm$ 0.73}}& \multicolumn{1}{c|}{99.30$\pm$ 0.44}\\ \hline
\end{tabularx}%
}
\end{table*}
\begin{figure*}[!t]
\centering
\subfigure[$m_{1}$]
{
    \includegraphics[width=0.23\linewidth]{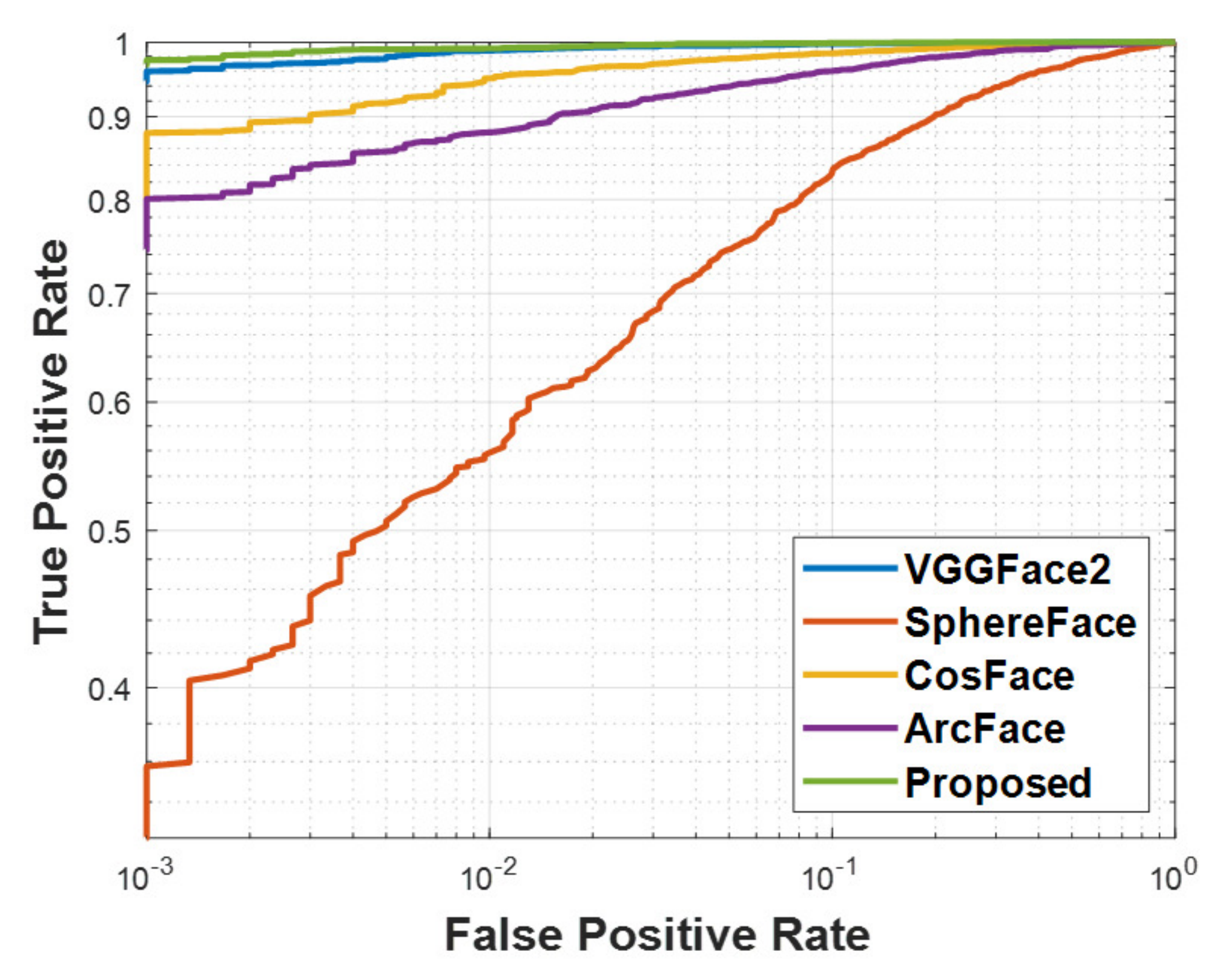}
}
\subfigure[$m_{2}$]
{
    \includegraphics[width=0.23\linewidth]{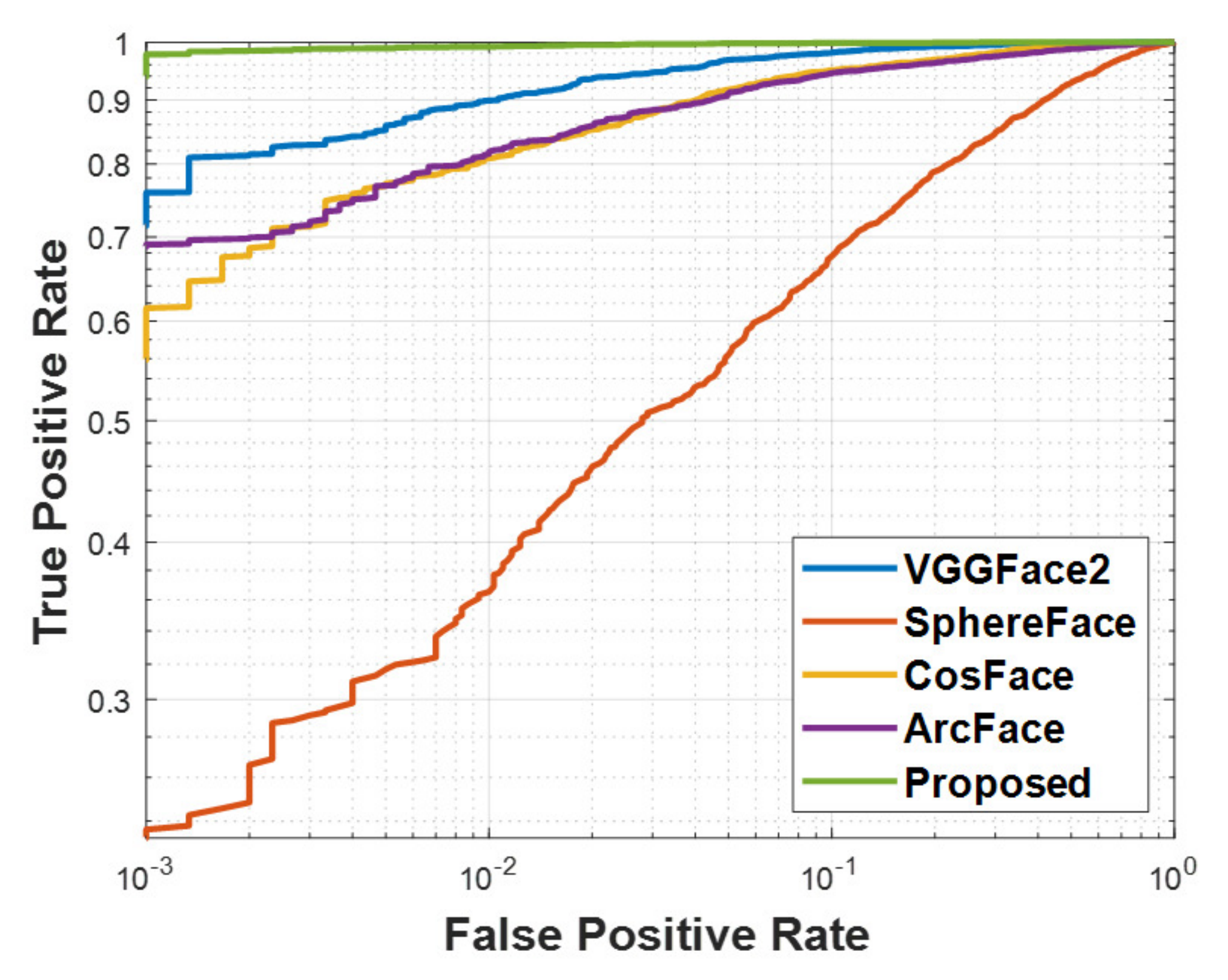}
}
\subfigure[$m_{3}$]
{
    \includegraphics[width=0.23\linewidth]{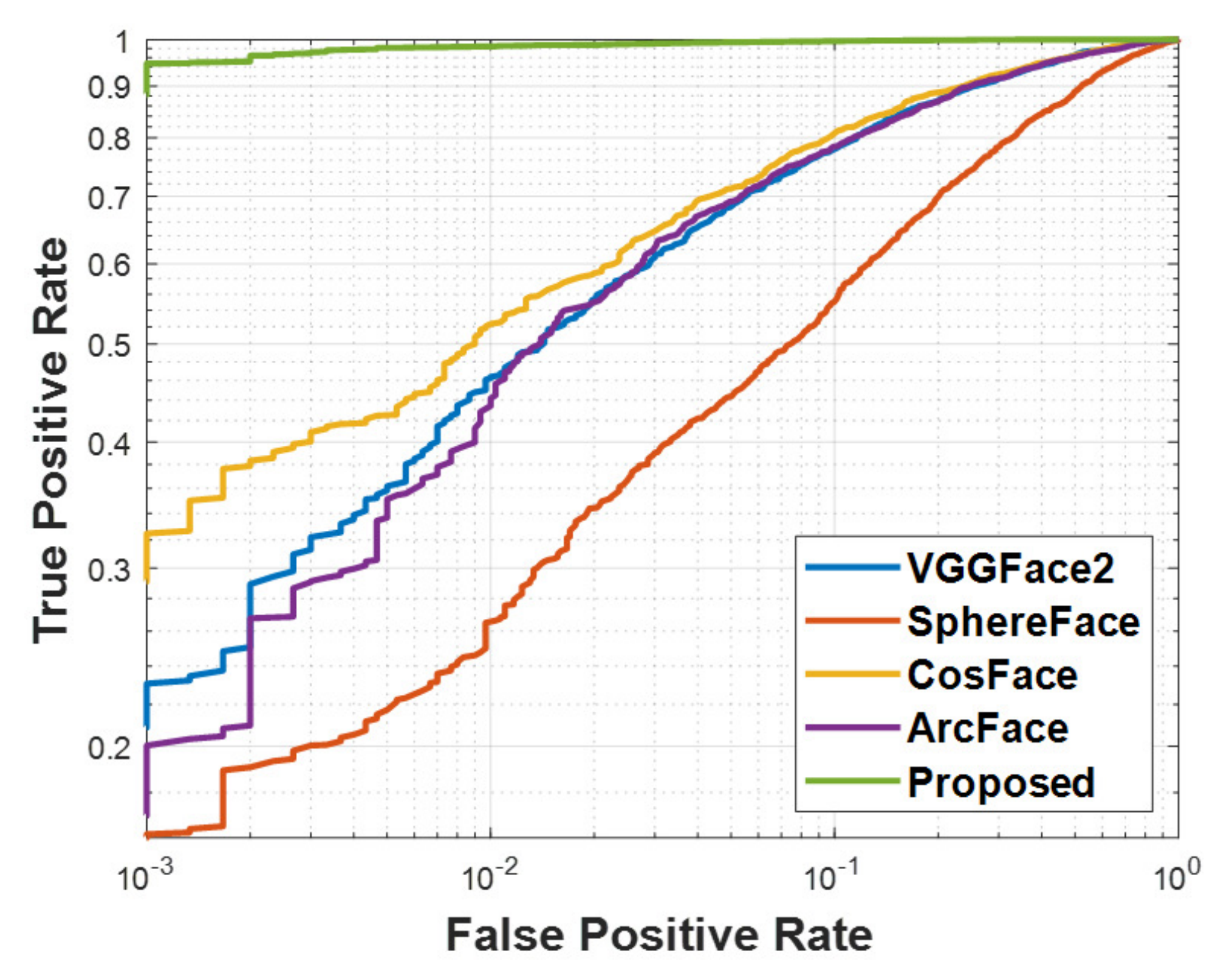}
}
\subfigure[$m_{4}$]
{
    \includegraphics[width=0.23\linewidth]{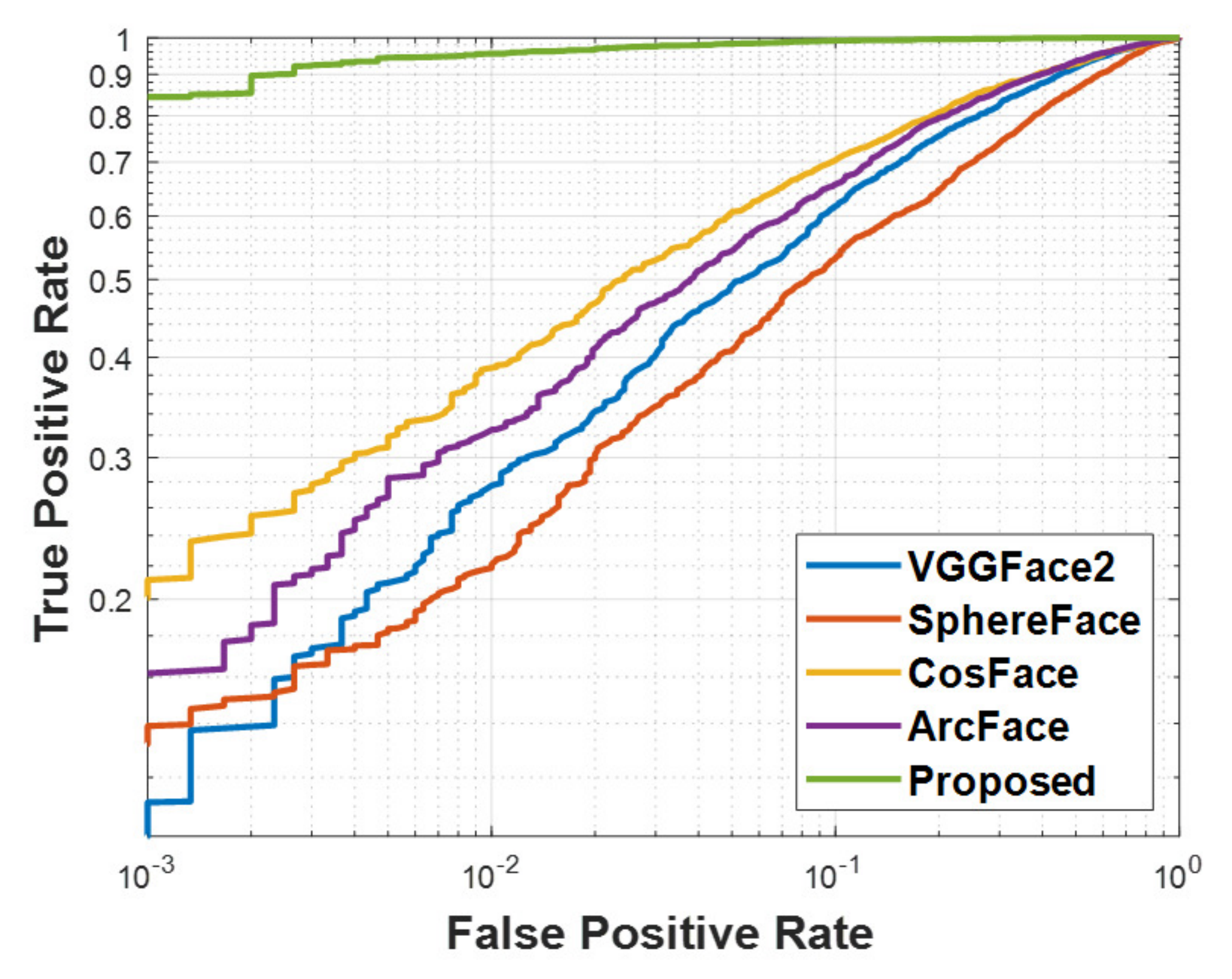}
}\\
\subfigure[$m_{5}$]
{
    \includegraphics[width=0.23\linewidth]{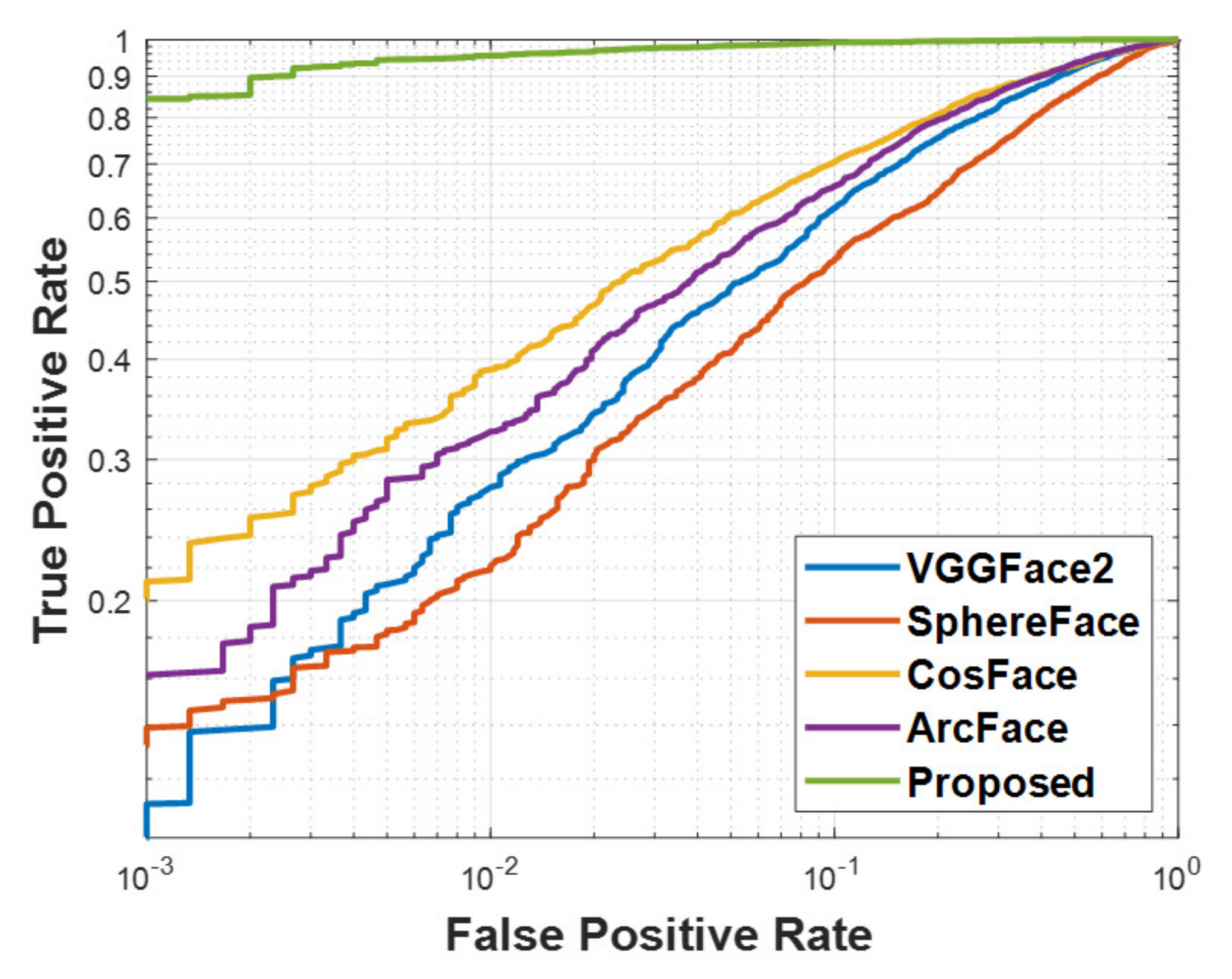}
}
\subfigure[$m_{6}$]
{
    \includegraphics[width=0.23\linewidth]{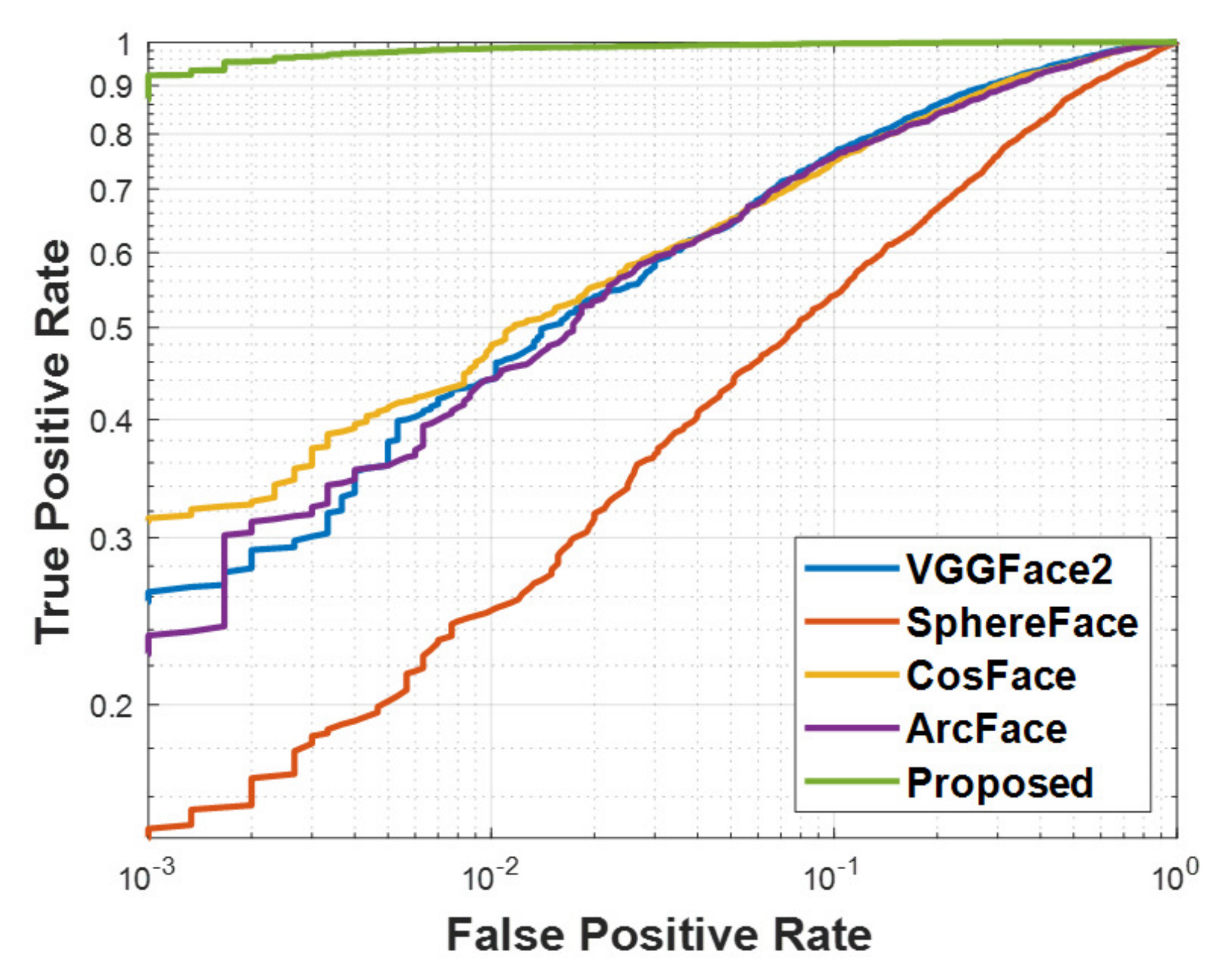}
}
\subfigure[$m_{7}$]
{
    \includegraphics[width=0.23\linewidth]{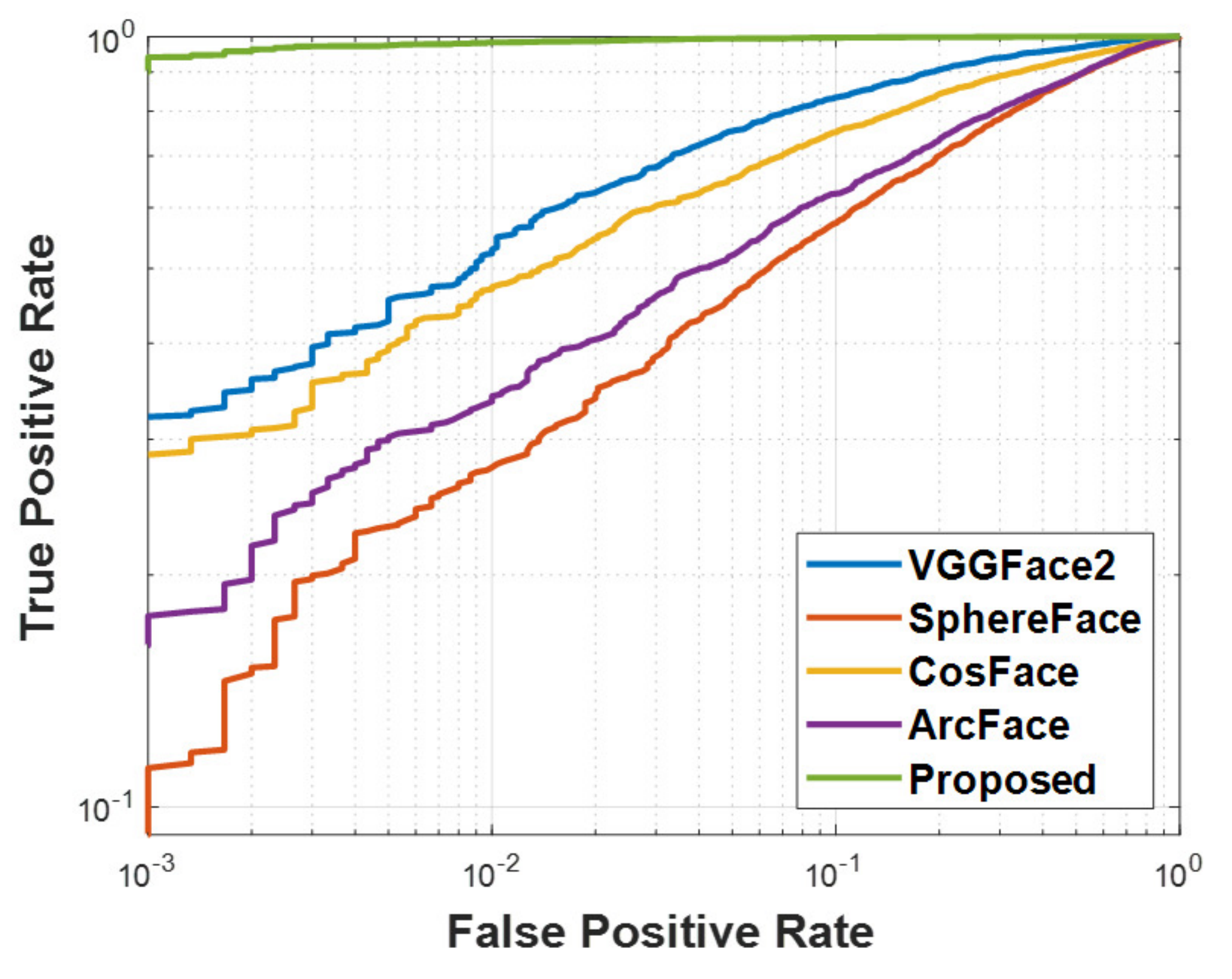}
}
\subfigure[Optimal Alignment]
{
    \includegraphics[width=0.23\linewidth]{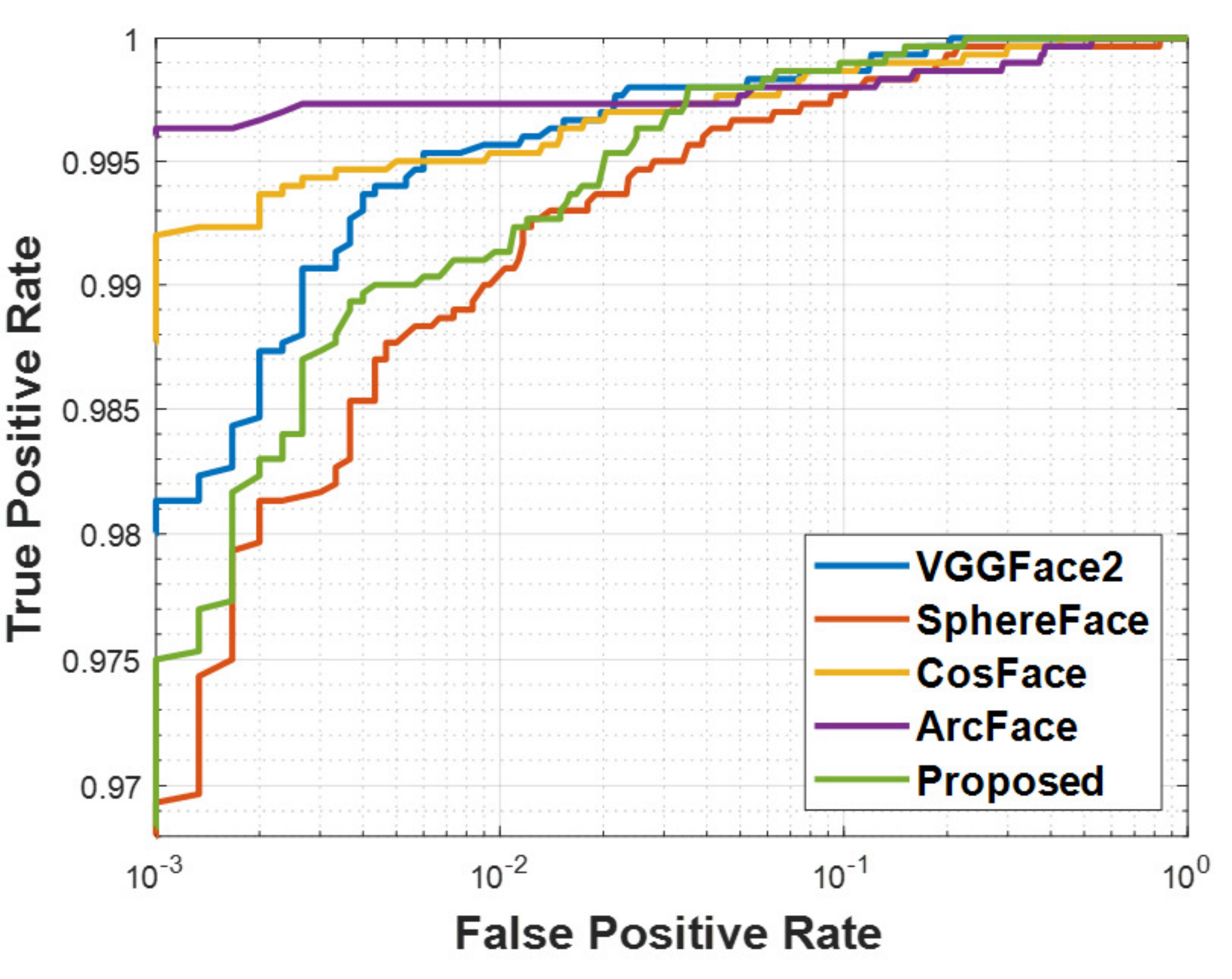}
}\\
\caption{Face verification ROC curves for the LFW dataset depending on the margin parameters. Best viewed in color.}
\label{fig_roc}
\end{figure*}

\textbf{Training Settings.}
To train the proposed deep network, four NVIDIA Titan Xp GPUs with $12$GB GPU memory were used. 
The batch size for one epoch was set to $64$ samples (\textit{i.e.}, $64$ well-aligned and $64$ randomly cropped face images as a pair with the corresponding face ID labels). 
The learning rate ($\delta$) and momentum ($\tau$) parameters were set to $0.1$ and $0.9$ recommended in~\cite{cao2018vggface2, he2016deep}, respectively.
Specifically, since the pre-trained model for the proposed network is not available and learning from scratch is required, a relatively large learning rate was set.
The learning rate for training was decayed by the factor of $0.1$ for every $30$ epochs. 
We set the total number of epochs to $100$. 
Note that the face shape estimation network was frozen to provide stable keypoint heatmaps in training. 
The image size to be decoded in the face image decoder was set to $112\times112$ pixels to reduce the memory overhead.
And, the standard deviation in the Gaussian blurring for generating heatmaps from the keypoint coordinates was set to $2$ (\textit{i.e.}, $\sigma=2$). 
The hyper-parameters in Eq. (\ref{eq_loss}) were all set to 1 (\textit{i.e.}, $\alpha=\beta=\gamma=1$).
Finally, the proposed method was optimized by a mini-batch gradient descent algorithm. 
The data preprocessing and training settings are summarized in Table~\ref{tab_set}.

\subsection{Effect of the Face Misalignment for Deep FR}\label{sec_exp1}
In this experiment, we first evaluated the LFW dataset in order to investigate the effect of the face misalignment for deep FR algorithms by adjusting the degree of the face misalignment. 
To adjust the face alignment for testing face images, we introduced margin parameters $m=\left(m_{x_{1}},m_{x_{2}},m_{y_{1}},m_{y_{2}}\right)$, where each element denotes a ratio controlling the margin for each point of a bounding box. 
Here, the face image's bounding box was defined by the two points obtained from a face detector: the left top point $\mathbf{x}_{1}=\left(x_{1},y_{1}\right)^\top$ and the right bottom point $\mathbf{x}_{2}=\left(x_{2},y_{2}\right)^\top$. 
Given the face image's bounding box, the bounding box for the misaligned face image (\textit{i.e.}, $\mathbf{x}_{1}'=\left(x_{1}',y_{1}'\right)^\top$ and $\mathbf{x}_{2}'=\left(x_{2}',y_{2}'\right)^\top$) was obtained by the following equation: 
\begin{equation}
\begin{pmatrix}
x_{1}'\\ 
x_{2}'
\end{pmatrix}
=\begin{pmatrix}
1+0.5m_{x_{1}} & -0.5m_{x_{1}}\\ 
-0.5m_{x_{2}} & 1+0.5m_{x_{2}} 
\end{pmatrix}
\begin{pmatrix}
x_{1}\\ 
x_{2}
\end{pmatrix},
\label{eq_trans1}
\end{equation}
\begin{equation}
\begin{pmatrix}
y_{1}'\\ 
y_{2}'
\end{pmatrix}
=\begin{pmatrix}
1+0.5m_{y_{1}} & -0.5m_{y_{1}}\\ 
-0.5m_{y_{2}} & 1+0.5m_{y_{2}} 
\end{pmatrix}
\begin{pmatrix}
y_{1}\\ 
y_{2}
\end{pmatrix}.
\label{eq_trans2}
\end{equation}
For the evaluation, we generated the seven types of misaligned face images: $m_{1}=\left(0.50,0.50,0.50,0.50\right)$, $m_{2}=\left(1.00,1.00,1.00,1.00\right)$, $m_{3}=\left(1.50,1.50,1.50,1.50\right)$,
$m_{4}=\left(2.00,2.00,2.00,2.00\right)$, $m_{5}=\left(2.50,2.50,2.50,2.50\right)$, $m_{6}=\left(1.25, 0.70, 1.75, 2.15\right)$, and $m_{7}=\left(0.33, 2.13, 2.17, 2.34\right)$. 
The margin parameters from $m_{1}$ to $m_{5}$ adjusted the face bounding boxes with the same ratio for all directions (left/right/top/bottom). 
The $m_{6}$ and $m_{7}$ adjusted the face bounding boxes differently for all directions, where each element was randomly selected. 
Fig.~\ref{fig_align} shows the example face images depending on the margin parameters used in this experiment.
\begin{table*}[!t]
\begin{center}
\caption{Face recognition accuracy (\%) for the LFW and CALFW datasets under the different types of alignment conditions.}
\label{tab_random}
\vspace{-0.3cm}
\begin{tabular}{|c||ccc|ccc|}
\hline
            & \multicolumn{3}{c|}{LFW} & \multicolumn{3}{c|}{CALFW} \\ \cline{2-7} 
 &  
  \begin{tabular}[c]{@{}c@{}}Optimal\\ Alignment\end{tabular} &
  \begin{tabular}[c]{@{}c@{}}Random\\ Alignment\end{tabular} &
  \begin{tabular}[c]{@{}c@{}}Without\\ Detection\end{tabular} &
  \begin{tabular}[c]{@{}c@{}}Optimal\\ Alignment\end{tabular} &
  \begin{tabular}[c]{@{}c@{}}Random\\ Alignment\end{tabular} &
  \begin{tabular}[c]{@{}c@{}}Without\\ Detection\end{tabular} \\ \Xhline{3\arrayrulewidth}
SphereFace~\cite{liu2017sphereface}          & 99.10$\pm$ 0.38    & 56.87$\pm$ 2.03    & 72.27$\pm$ 2.52   & 89.53$\pm$ 0.52      & 52.78$\pm$ 2.33      & 56.72$\pm$ 2.47      \\
CosFace~\cite{wang2018cosface}            & 99.52$\pm$ 0.30    & 64.60$\pm$ 1.28    & 80.12$\pm$ 1.67   & 90.62$\pm$ 0.41      & 58.32$\pm$ 1.56      & 61.83$\pm$ 1.63      \\
VGGFace2~\cite{cao2018vggface2}          & 99.08$\pm$ 0.54    & 85.71$\pm$ 1.85    & 77.87$\pm$ 1.82   & 87.75$\pm$ 0.70     & 63.80$\pm$ 1.75      & 59.57$\pm$ 1.81      \\
ArcFace~\cite{deng2019arcface}            & \textbf{99.72$\pm$ 0.19}    & 64.09$\pm$ 1.81    & 79.00$\pm$ 2.03   & \textbf{93.63$\pm$ 0.22}      & 59.93$\pm$ 1.80      & 63.85$\pm$ 1.94      \\ \hline
Proposed           & 99.30$\pm$ 0.44    & \textbf{97.46$\pm$ 0.95}    & \textbf{97.17$\pm$ 0.98}   & 90.61$\pm$ 0.51      & \textbf{89.14$\pm$ 0.82}      & \textbf{88.45$\pm$ 0.88}       \\ \hline
\end{tabular}
\end{center}
\end{table*}
\begin{figure*}[!t]
\centering
\subfigure[LFW]
{
    \includegraphics[width=0.48\linewidth]{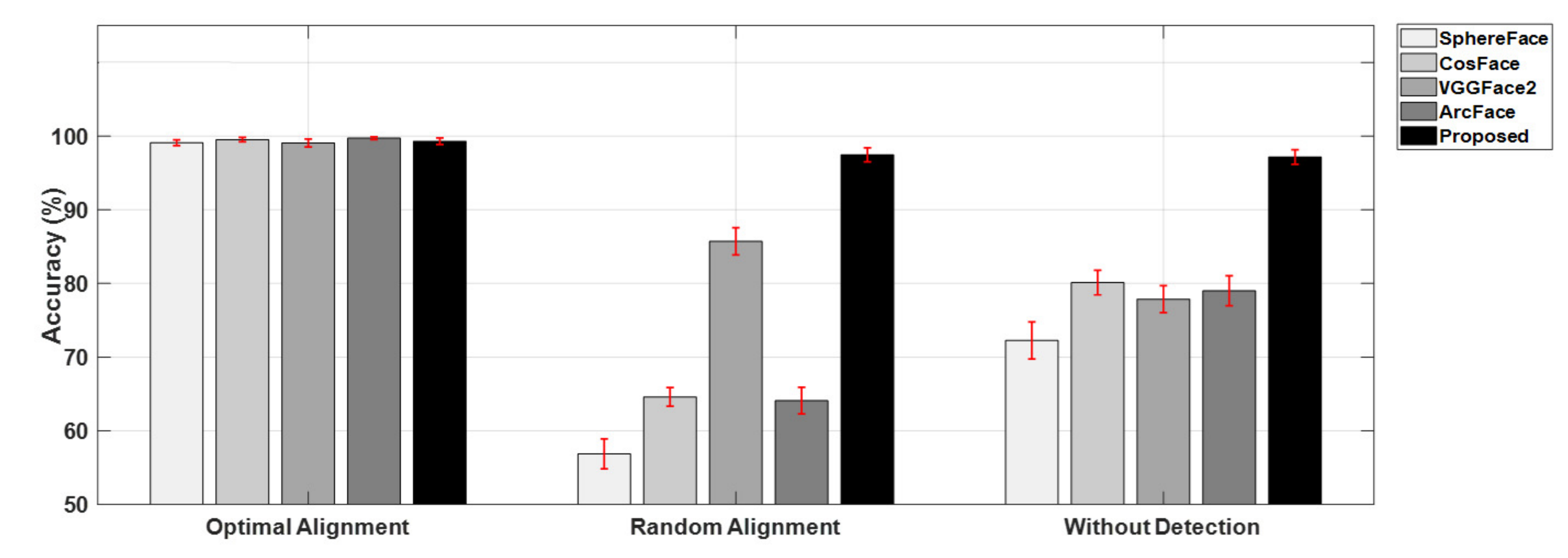}
}
\subfigure[CALFW]
{
    \includegraphics[width=0.48\linewidth]{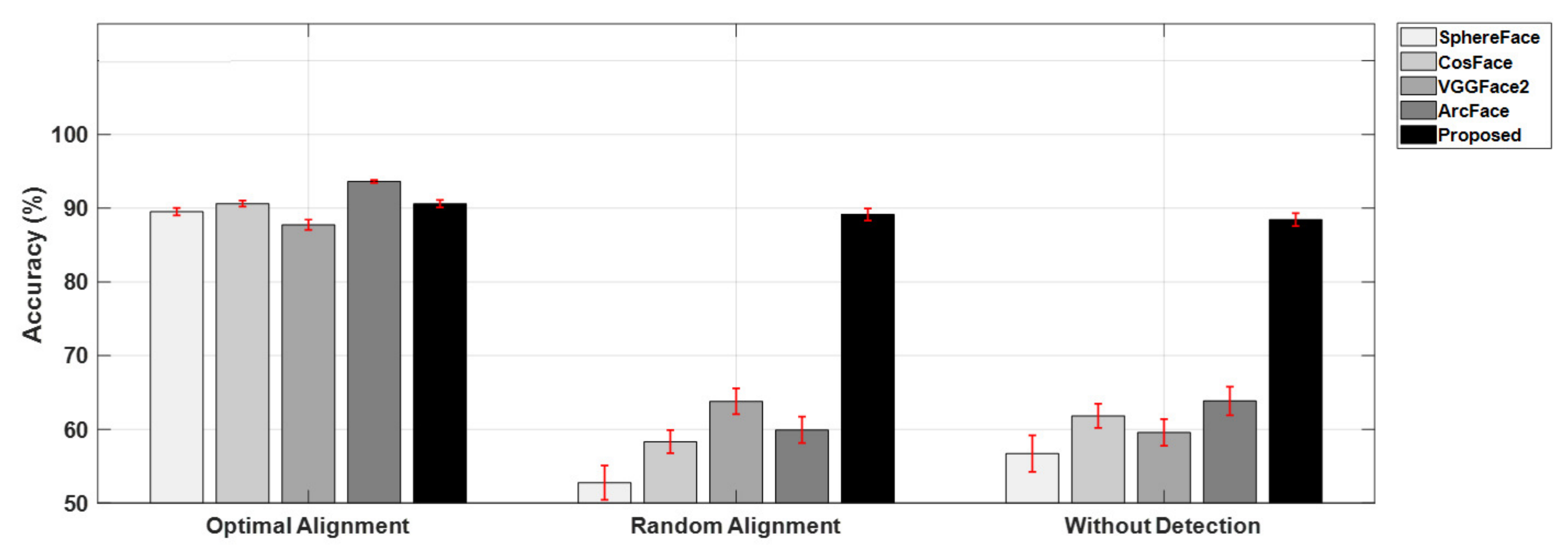}
}
\vspace{-0.3cm}
\caption{Error bar graphs for the face recognition accuracy (\%) for the LFW and CALFW datasets presented in Table~\ref{tab_random}. }
\label{fig_errbar}
\end{figure*}
By increasing the margin parameter, we obtain the face image similar to the original input face image before cropping. 
For the evaluation of the robustness to the face misalignment, the state-of-the-art deep FR algorithms were compared: 1) SphereFace~\cite{liu2017sphereface} ($64$-layer CNN with a residual unit trained with CASIA WebFace~\cite{yi2014learning}), 2) CosFace~\cite{wang2018cosface} ($64$-layer CNN with a residual unit trained with CASIA WebFace), 3) VGGFace2~\cite{cao2018vggface2} ($50$-layer CNN with a residual unit trained with VGGFace2 dataset~\cite{cao2018vggface2}), and 4) ArcFace~\cite{deng2019arcface} ($101$-layer CNN with an improved residual unit trained with MS-Celeb-1M~\cite{guo2016ms}).
Note that the SphereFace, CosFace, and ArcFace used basically horizontal flipping as data augmentation for training the deep networks~\cite{liu2017sphereface, wang2018cosface, deng2019arcface}. 
In contrast, the VGGFace2 algorithm adopted the over-sampling strategy~\cite{cao2018vggface2} that extends the face bounding box, then randomly crops as the data augmentation.
Based on the publicly available deep FR models, we measured the accuracy in our machine under the same conditions. 

Table~\ref{tab_lfw2} shows the face verification performance for the LFW dataset depending on the margin parameters.
And Fig.~\ref{fig_roc} shows the face verification receiver operating characteristic (ROC) curves for the LFW dataset depending on the margin parameters.
The performance (\textit{i.e.}, average performance and standard deviation) was measured based on the splits for 10-fold cross-validation in~\cite{huang2008labeled}.
The standard deviation can be thought of as a measure of how stable a recognition performance is for repeated experiments, and a small standard deviation value is preferred.
Note that the ``Optimal Alignment'' means the face verification results accompanied by the optimal alignment process used in each deep FR algorithm. 
For the three deep FR algorithms (SphereFace, CosFace, and ArcFace), a face image is aligned based on the facial keypoints, then crop the face images with $96\times112$ pixels (for SphereFace), $112\times112$ pixels (for both CosFace and ArcFace). The VGGFace2 algorithm crops face images based on the bounding boxes without facial keypoint estimation. 
Even if the previous deep FR algorithms showed the high performance for the face images obtained from the optimal face alignment, the accuracy was significantly degraded when changing the type of face alignment. 
It shows that the performance of the existing methods becomes highly sensitive to face alignment. 
Furthermore, standard deviation values become large (\textit{i.e.}, decreased stability) as the type of face alignment changes.
In the case of the SphereFace, the degradation of the performance was significant. 
This is mainly because the input face image of the SphereFace has a different ratio for height and width. 
By changing the margin parameter, the ratio of a face image is considerably degraded.
In the case of the VGGFace2, it could be tolerant as the margin parameter varies thanks to the over-sampling strategy in Table~\ref{tab_lfw2}.
However, the accuracy for the VGGFace2 algorithm was also degraded according to the increase of the margin parameter despite the advantage of the over-sampling. 
In contrast, the proposed method showed robust performance with a relatively small standard deviation to the change of the margin parameters.
Even more, we observe that the proposed method shows comparable accuracy though the face image is cropped with high margin values including backgrounds. 
In other words, the basic data augmentation strategy like the over-sampling can be used as one of the solutions that resolve the face misalignment problem, however, it would not be an optimal solution.

\subsection{Robustness under Random Alignment for Deep FR}\label{sec_exp2}
In the conventional FR pipeline, there are errors and uncertainties caused by face detection and face alignment. 
These uncertainties of the face detector and face keypoint detector make the face images cropped with a different type of alignment. 
In this experiment, we changed the margin parameters randomly instead of a specific margin parameter as presented in Section~\ref{sec_exp1} (we denote the case as ``Random Alignment''). 
Here, the margin parameters were randomly generated from $0$ to $3$ by a uniform distribution (\textit{i.e.}, $\mathcal{U}\left(0,3\right)$), which were applied to four elements of the margin parameter independently. 
Additionally, we compared the accuracy under the original test image as it is (without a face detection, which is denoted as ``Without Detection'') as the extreme case. 
Note that the accuracy of ``Optimal Alignment'' for the proposed method was selected by the maximum performance between $m_{1}$ and $m_{7}$. 
Table~\ref{tab_random} shows the face verification performance under three alignment conditions for the LFW and CALFW datasets with error bar graphs in Fig.~\ref{fig_errbar}. 
Without the optimal face alignment, the accuracies for the previous methods were significantly degraded as observed in Section~\ref{sec_exp1}. 
In particular, we observe that the accuracy for ``Random Alignment'' is worse than that of ``Without Detection''. 
This is mainly caused by the mismatches between test face images in a pair. 
In contrast, the proposed method showed robust performance with a relatively small standard deviation even under different types of face misalignment conditions.
\begin{figure}[!t]
\centering
\includegraphics[width=0.8\linewidth]{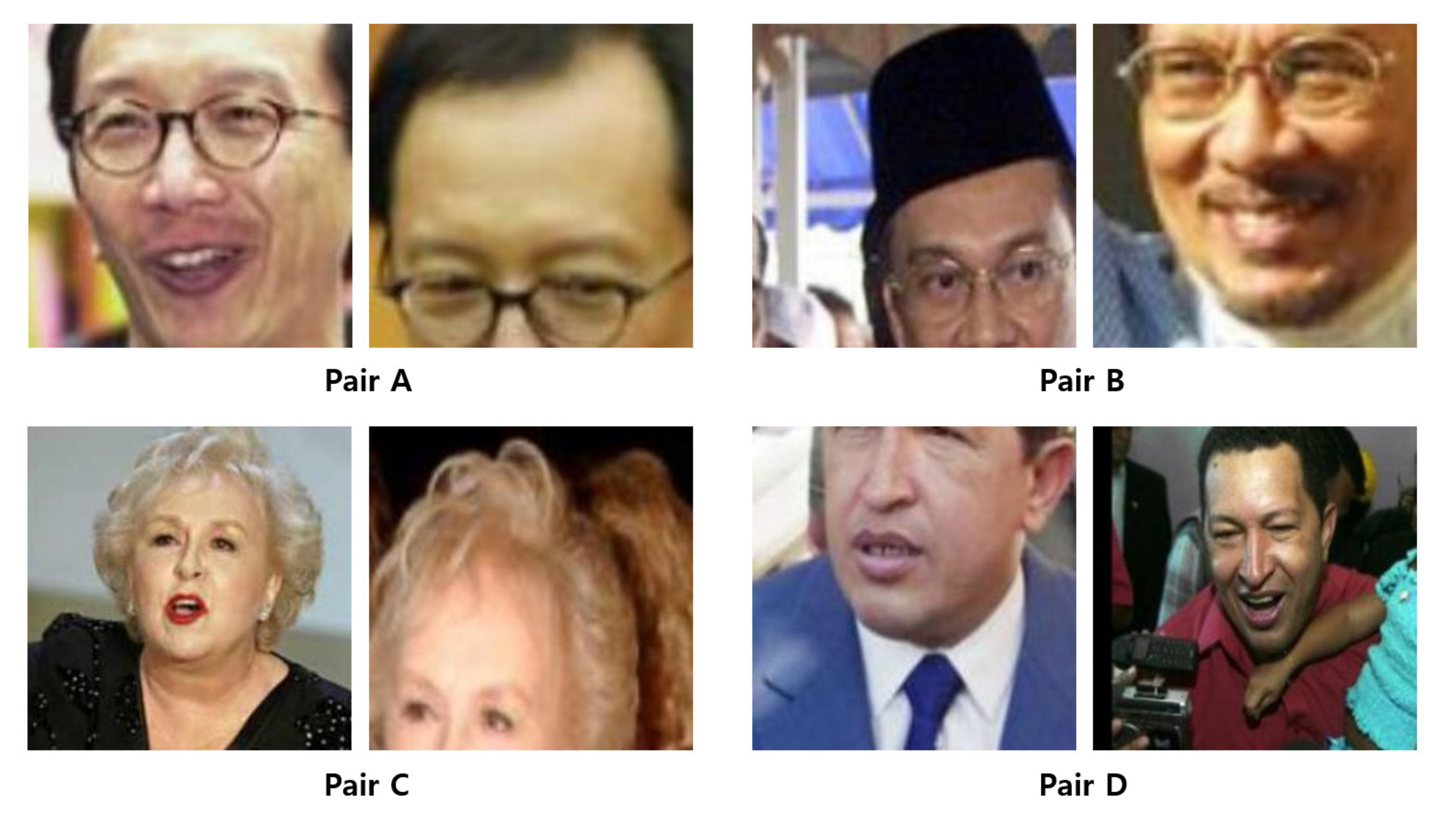}
\begin{center}
\end{center}
\vspace{-1.0cm}
   \caption{Example face images for the failure cases of the proposed method.}
\label{fig_failure}
\vspace{-0.5cm}
\end{figure}
Of course, we can observe few failure cases for the proposed method: 1) when the face bounding box contains large amounts of background information other than the face area as it is extended by the margin parameter, 2) when the partial face not including all face components (eyes, nose, or mouth) by the random alignment is used for testing. Fig.~\ref{fig_failure} shows the example failure cases for the proposed method. 
These failure cases were included in order to test the proposed method under extreme conditions in the experiment. 
However, we can observe that the proposed method shows robustness to extreme conditions compared to the previous works.

\begin{table}[!t]
\caption{LFW face verification accuracy (\%) depending on the combination of the loss functions: $\mathcal{L}_{cls}$ (loss for classification), $\mathcal{L}_{pa}$ (loss for pixel alignment), and $\mathcal{L}_{fa}$ (loss for feature alignment).}
\label{tab_loss}
\centering
\vspace{-0.3cm}
\resizebox{\linewidth}{!}{%
\begin{tabular}{|l||c|c|}
\hline
\multirow{2}{*}{Loss functions for training} & \multicolumn{2}{c|}{LFW}             \\ \cline{2-3}
& Random Alignment & Without Detection \\ 
\Xhline{3\arrayrulewidth}
$\mathcal{L}_{cls}$              & 84.01$\pm$ 1.43            & 74.07$\pm$ 1.62             \\
$\mathcal{L}_{cls}+\mathcal{L}_{pa}$              & 92.65$\pm$ 1.14            & 93.05$\pm$ 1.22             \\
$\mathcal{L}_{cls}+\mathcal{L}_{fa}$              & 95.32$\pm$ 0.83            & 95.15$\pm$ 0.91             \\
$\mathcal{L}_{cls}+\mathcal{L}_{pa}+\mathcal{L}_{fa}$             & 97.46$\pm$ 0.95            & \textbf{97.18$\pm$ 0.98}             \\
$\mathcal{L}_{cls}+2\mathcal{L}_{pa}+\mathcal{L}_{fa}$                         & 97.23$\pm$ 0.92            & 97.00$\pm$ 0.87             \\
$\mathcal{L}_{cls}+\mathcal{L}_{pa}+2\mathcal{L}_{fa}$            & \textbf{97.62$\pm$ 0.88}            & 97.08$\pm$ 0.94              \\ \hline
\end{tabular}%
}
\end{table}

\subsection{Exploratory Experiments}\label{sec_exp3}
In this section, we investigate the effectiveness of the proposed method via exploratory experiments: 1) effect of the loss functions, and 2) effect of the face shape estimation network.

\textbf{Effect of the Loss Functions.} First, we reported the accuracy depending on the loss functions to investigate the effect of the three different loss functions. 
According to Table~\ref{tab_loss}, when the classification loss ($\mathcal{L}_{cls}$) was only considered, the performance was sensitive to the face misalignment and showed similar accuracy to VGGFace2.
In contrast, when the pixel alignment loss ($\mathcal{L}_{pa}$) reflecting the reconstruction loss between the well-aligned and randomly cropped face images was additionally used, the accuracy was improved. 
We observe that the additional shape-related information by the pixel alignment loss gives a positive effect to train the deep network even though there is no direct guidance from the feature alignment process. 
Additionally, when the feature alignment loss ($\mathcal{L}_{fa}$) was considered with the $\mathcal{L}_{cls}$, the accuracy was more enhanced thanks to the direct guidance of the shape-aggregated feature. 
Finally, we achieved the best performance when all loss functions were accommodated. 
Also, we evaluated the accuracy for the three-loss functions with different hyper-parameters that control the balance for the total loss function. 
As can be seen in Table~\ref{tab_loss}, the additional performance gain could be obtained when the feature alignment loss was more weighted than the pixel alignment loss. 
It turns out that the feature alignment loss is much more important than the pixel alignment loss. 
In conclusion, this experiment demonstrates that the three-loss functions are complementary to each other even if the accuracy can vary by the hyper-parameters selection. 
Combining all functions is highly effective for the purpose of FR robust to face misalignment.

\begin{table}[!t]
\begin{center}
\caption{LFW face verification accuracy (\%) depending on the face shape estimation network, where the keypoint estimation performance (\%) by the face shape estimation network was measured AUC @ 8\% NME~\cite{wang2019adaptive} with 300W private dataset~\cite{sagonas2013300}.}
\label{tab_shape}
\vspace{-0.5cm}
\resizebox{\linewidth}{!}{%
\begin{tabular}{|c||c|c|c|c|c|c|}
\hline
\multirow{3}{*}{\begin{tabular}[c]{@{}c@{}} Face Shape Estimation \\ Network (FSEN)\end{tabular}}&
\multirow{3}{*}{\begin{tabular}[c]{@{}c@{}} Keypoint \\ Estimation \\ Performance\end{tabular}} &
\multicolumn{2}{c|}{\begin{tabular}[c]{@{}c@{}} LFW \\ (1:1 Verification) \end{tabular}}\\ \cline{3-4} 
                    &                           & 
                  \begin{tabular}[c]{@{}c@{}} Optimal \\ Alignment \end{tabular}&
                  \begin{tabular}[c]{@{}c@{}} Random \\ Alignment \end{tabular}\\ \Xhline{3\arrayrulewidth}
Proposed w/o FSEN & - & 99.04$\pm$0.71 &  95.25$\pm$0.60\\ 
Proposed w/ FSEN (FAN$_{\epsilon=4}$)    & 38.34 &  98.97$\pm$0.54 &  96.78$\pm$0.56\\ 
Proposed w/ FSEN (FAN$_{\epsilon=2}$)    &  45.49 & 99.23$\pm$0.65 &  \textbf{97.46$\pm$0.58}\\ 
Proposed w/ FSEN (FAN$_{\epsilon=0}$)  &  \textbf{48.38} & \textbf{99.30$\pm$0.44} &  \textbf{97.46$\pm$0.95}\\ \hline
\end{tabular}%
}
\end{center}
\end{table}
\textbf{Effect of the Face Shape Estimation Network.} In order to investigate the contribution of the face shape information, we conducted the ablation experiments for two cases: 1) when the face shape estimation network (FSEN) was removed, 2) when the different face shape estimation network with different keypoint estimation performance was used. 
For the first case (i.e., Proposed w/o FSEN in Table~\ref{tab_shape}), the experimental settings were set to the same as the previous experimental conditions, and only the existence of the face shape estimation network is different. 
When the face shape estimation network is removed, the stacked feature map becomes the same as the face appearance feature map ($\mathbf{h}_{i}^{F}$) because the face shape heatmap is not available. 
According to Table~\ref{tab_shape}, the accuracy dropped by about 2\% when the FSEN was removed under the random alignment. 
Considering that the 1\% gain of the recognition performance for the LFW dataset is significant, the FSEN in the proposed method helps improve the recognition performance with robustness. 
In other words, the explicit face shape prior as well as the well-aligned face image achieves a synergy effect for improving the accuracy to the face misalignment. 
To deal with the second case, we added a Gaussian random noise for the result of the landmark prediction by the FAN. Let us denote the originally estimated landmark by the FAN as $k_{x}$ and $k_{y}$. 
Then, the perturbed landmarks by the noise were computed by $\tilde{k}_{x} = k_{x} + n_{x}$ and $\tilde{k}_{y} = k_{y} + n_{y}$, where $n_{x}$ and $n_{y}$ follow a Gaussian distribution, i.e., $n_{x}, n_{y}\sim \mathcal{N}\left(0, \epsilon^{2}\right)$.
In our experiment, we set the standard deviation ($\epsilon$) to 2 and 4 (i.e., Proposed w/ FSEN (FAN$_{\epsilon=2}$) and Proposed w/ FSEN (FAN$_{\epsilon=4}$) in Table~\ref{tab_shape}).
Note that the proposed method without the noise perturbation in the FSEN is denoted as Proposed w/ FSEN (FAN$_{\epsilon=0}$). 
The FSEN performance was measured by the area under the curve (AUC) at 8\% normalized mean error (NME)~\cite{wang2019adaptive} based on the bounding box size normalization~\mbox{\cite{bulat2017far}} with 300W private dataset~\cite{sagonas2013300} including 300 indoors and 300 outdoor faces.
As shown in Table~\ref{tab_shape}, the accuracy did not change sensitively depending on the face keypoint estimation accuracy. 
This is mainly because the proposed method utilized the face keypoint heatmap instead of the localization point. Note that the heatmap is for emphasizing the importance of the surrounding areas around the keypoints~\cite{Khorramshahi_2019_ICCV}, which can be tolerable to some extent of noise.

\begin{table}[!t]
\caption{CPLFW face verification accuracy (\%) with or without metadata provided in the dataset. ``Difference'' (\%p) denotes the difference in average accuracy for the two cases.}
\label{tab_pose}
\centering
\vspace{-0.3cm}
\resizebox{\linewidth}{!}{%
\begin{tabular}{|c||c|c|c|}
\hline
\multirow{2}{*}{} & \multicolumn{3}{c|}{\begin{tabular}[c]{@{}c@{}} CPLFW \\ (1:1 Verification) \end{tabular}}             \\ \cline{2-4}
& \begin{tabular}[c]{@{}c@{}} With \\ metadata \end{tabular} & \begin{tabular}[c]{@{}c@{}} Without \\ metadata \end{tabular} &Difference \\ 
\Xhline{3\arrayrulewidth}
SphereFace              & 79.33$\pm$ 2.45            & 75.97$\pm$ 3.42  & 3.37 $\downarrow$           \\
CosFace              & 85.58$\pm$ 2.09 & 79.45$\pm$ 2.30     & 6.13 $\downarrow$        \\
VGGFace2 & 83.85$\pm$ 2.07            & 79.77$\pm$ 2.96       & 4.08 $\downarrow$      \\
ArcFace             & \textbf{88.10$\pm$ 2.58}            & 83.17$\pm$ 2.53  & 4.93 $\downarrow$           \\ \hline
Proposed                         & 85.47$\pm$ 2.59            & \textbf{85.40$\pm$ 2.59}     & \textbf{0.07 $\downarrow$}        \\ \hline
\end{tabular}%
}
\end{table}
\begin{figure}[!t]
\centering
\subfigure[False Negatives]
{
    \includegraphics[width=1.0\linewidth]{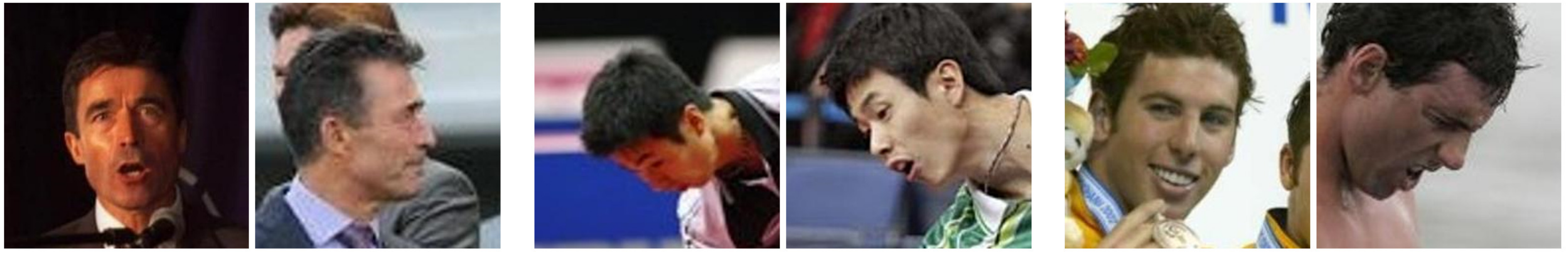}
}\\
\subfigure[False Positives]
{
    \includegraphics[width=1.0\linewidth]{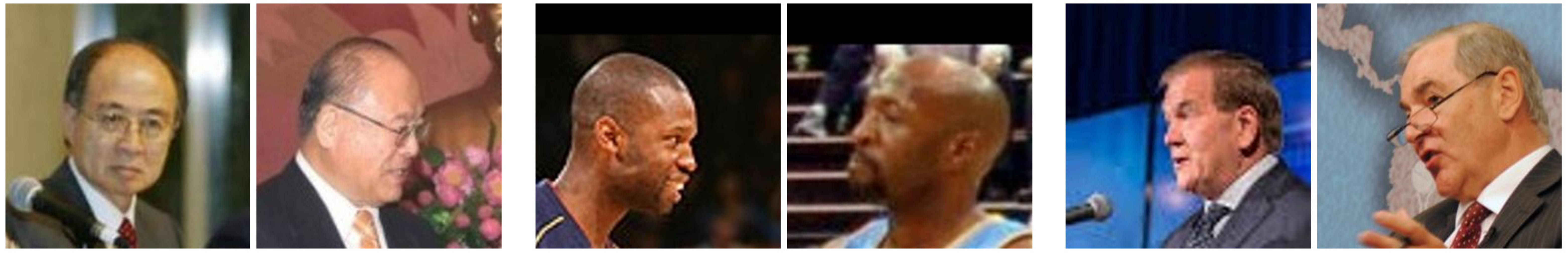}
}\\
\caption{Example face images for the failure case of the proposed method on the CPLFW dataset.}
\label{fig_cplfw}
\end{figure}
\textbf{Experiments with Large Pose Variations.} 
To further investigate the recognition performance under large pose variations where explicit face alignment could not be well addressed, we performed experiments on the CPLFW dataset~\cite{zheng2018cross}. 
As discussed in Section~\ref{sec:expset}, the CPLFW dataset is comprised of 3,000 positive and negative pairs (\textit{i.e.}, 6,000 pairs), where the positive pairs with pose differences were collected to add pose variations to intra-class variance. 
Note that the range of pose variations in the CPLFW is known to be between $-90^{\circ}$ and $+90^{\circ}$ in yaw.
With the CPLFW dataset, we compared the experimental results for the two cases: 1) ``With metadata'' that uses the bounding box and landmark information (\textit{i.e.}, metadata) provided by the CPLFW dataset, 2) ``Without metadata'' that obtains the bounding box and landmark information from the MTCNN detector without the metadata. 
Table~\ref{tab_pose} shows the performance and performance differences for each case. 
We can see that the recognition performance for existing methods decreased by about 4-6\%p due to the failure of detecting the bounding box and landmarks. 
In contrast, the proposed method almost maintains the recognition performance (\textit{i.e.}, -0.07\%p difference). 
This result shows that the proposed method without the explicit face alignment is robust against face misalignment.
However, as can be seen in Fig.~\ref{fig_cplfw}, we can observe several failure cases for the proposed method, where most of the failure cases were those with very large pose variations. 
This is mainly because the range of pose variations in the VGGFace2 dataset used for training the proposed method is limited (\textit{i.e.}, most of the face images included in VGGFace2 have pose variations between $-40^{\circ}$ and $+40^{\circ}$ in yaw).

\begin{table*}[!t]
\begin{center}
\caption{The number of parameters for the proposed method, where the ``Train/Test'' means that the network parameters are used in both training and test. The ratio is computed based on the number of parameters for the total number of training parameters.}
\label{tab_params}
\vspace{-0.3cm}
\begin{tabular}{|c||c|r|c|c|}
\hline
                                          & Train/Test & Number of Parameters & Ratio \\ \Xhline{3\arrayrulewidth}
Facial Keypoint Estimation Network ($S$)  & Train      & 23,820,176                               & 67\% \\
Feature Extraction Network ($F$)          & Train/Test & 5,902,528                                & 17\%  \\
Face Image Decoder ($D$)                  & Train      & 555,712                                  & 2\%  \\
Fully Connected Layer ($\mathbf{Q}, \mathbf{b}$)            & Train      & 4,427,703                                & 12\% \\
Feature Map Embedding Layer ($\varphi$) & Train/Test & 525,312                                  & 1\%  \\
Channel Aggregation Layer ($\phi$)       & Train      & 526,848                                  & 1\% \\ \hline
\textbf{Total Number of Training Parameters} & Train & 35,758,279 & 100\% \\ \hline
\textbf{Total Number of Test Parameters}     & Test  & 6,427,840  & 18\%  \\ \hline
\end{tabular}
\end{center}
\end{table*}

\begin{figure*}[!t]
\centering
\subfigure[Input images for Grad-CAM visualization]
{
    \includegraphics[width=1.0\linewidth]{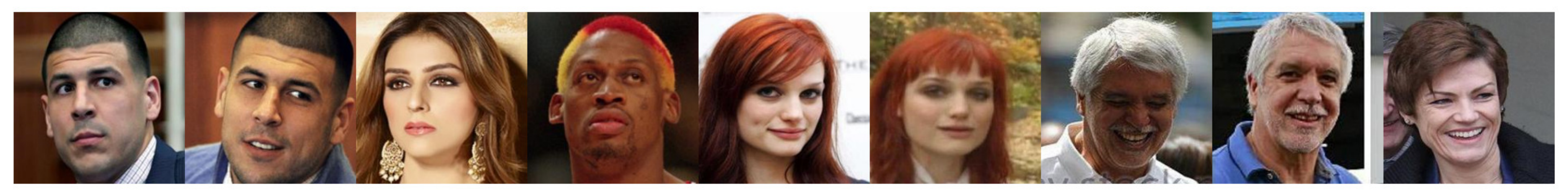}
}\\
\subfigure[Grad-CAM results without considering the face shape prior (Layered images)]
{
    \includegraphics[width=1.0\linewidth]{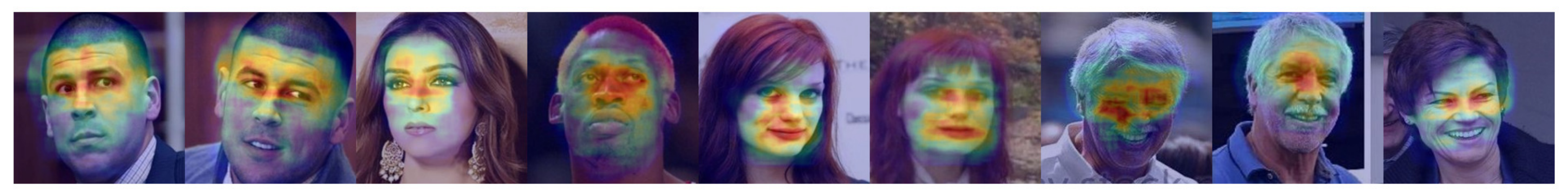}
}\\
\subfigure[Grad-CAM results without considering the face shape prior (Activations)]
{
    \includegraphics[width=1.0\linewidth]{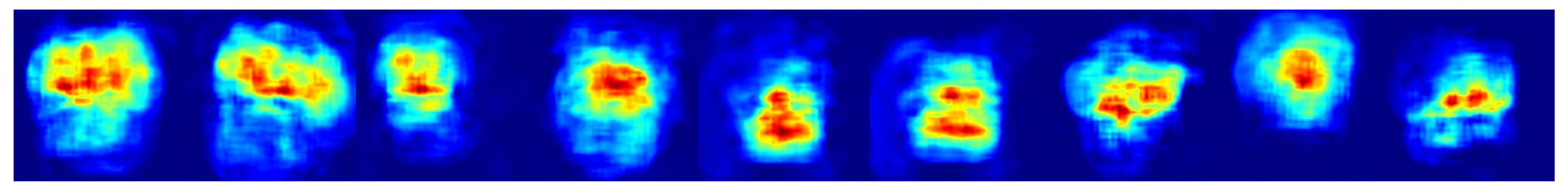}
    \centering
}\\
\subfigure[Grad-CAM results with considering the face shape prior (Layered images)]
{
    \includegraphics[width=1.0\linewidth]{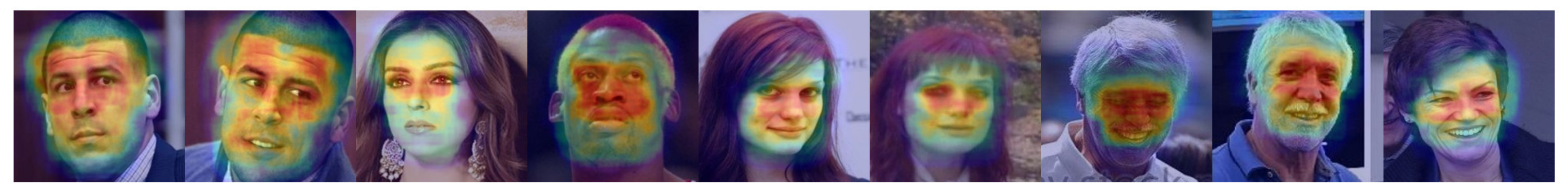}
}\\
\subfigure[Grad-CAM results with considering the face shape prior (Activations)]
{
    \includegraphics[width=1.0\linewidth]{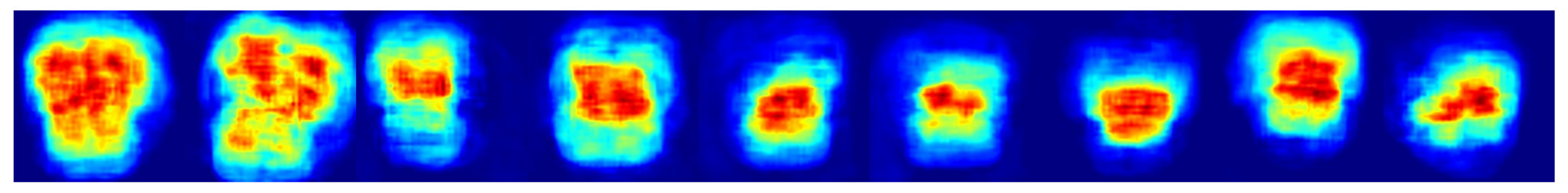}
}
\vspace{-0.3cm}
\caption{Visualization by the Grad-CAM. The top row shows input face images for the analysis. The second and third rows represent the results without considering the face shape prior (face feature extraction network only, \textit{i.e.}, the modified ResNet50). The fourth and fifth rows show the Grad-CAM results considering the face shape prior.}
\label{fig_gradcam}
\end{figure*}

\subsection{Number of Learning Parameters}\label{sec_exp4}
To validate the efficiency of the proposed method, we measured the number of parameters used for both training and testing stages in Table~\ref{tab_params}. 
Among the six different networks listed in Table~\ref{tab_arch}, the parameters for the facial keypoint estimation network occupied $67$\% of all parameters, which holds the majority number of parameters of the proposed deep network.
However, when testing a FR with our method, only two networks (\textit{i.e.}, feature extraction network and feature map embedding network) were used. 
In other words, $18$\% parameters (about $6$ million) from the total number of training parameters were only used. 
Therefore, since the proposed deep network only requires a similar number of parameters used in the previous deep FR models without additional computations, we can efficiently compute features robust to the face misalignment. 

\begin{table*}[!t]
\begin{center}
\caption{Comparison with the state-of-the-art deep FR and face alignment learning algorithms, where the FR performance for the MegaFace dataset was measured with the cleaned version of the dataset as suggested in~\cite{deng2019arcface}. $^\dagger$ denotes the MegaFace performance in~\mbox{\cite{zhong2017toward}} without filtering noise lists.}
\label{tab_sota}
\begin{tabular}{|c|c||c|c|c|c|}
\hline
& & {\begin{tabular}[c]{@{}c@{}} LFW \\ (1:1 Verification) \end{tabular}}              & {\begin{tabular}[c]{@{}c@{}} YTF \\ (1:1 Verification) \end{tabular}} & {\begin{tabular}[c]{@{}c@{}} MegaFace \\ (1:N Identification) \end{tabular}}\\ \Xhline{3\arrayrulewidth}
\multirow{10}{*}{Deep Face Recognition} & SphereFace~\cite{liu2017sphereface} &  99.42 &  95.00  & 72.73$^\dagger$\\ 
&  CosFace~\cite{wang2018cosface} &  99.73 &  97.60 & 82.72$^\dagger$\\ 
&  VGGFace2~\cite{cao2018vggface2}   &  99.08 &  97.30 & 94.17\\ 
& ArcFace~\cite{deng2019arcface} &  \textbf{99.82} & \textbf{98.02} & \textbf{98.35} \\
&UniformFace~\cite{duan2019uniformface} &  99.80 &  97.70 &  79.98$^\dagger$ \\
&MML~\cite{wei2020minimum} &  99.63 &  95.50 &  83.00$^\dagger$ \\
&DBM~\cite{cao2020domain} & 99.78 &  - &  96.35 \\
&CLMLE~\cite{huang2019deep} &  99.62 &  96.50 &  79.68$^\dagger$ \\
&CurricularFace~\cite{huang2020curricularface} &  99.80 &  - &  98.25 \\
&ARFace~\cite{zhang2021arface} &  99.62 &  97.54 &  96.40 \\
\hline\hline
\multirow{6}{*}{Face Alignment Learning} & Zhong \textit{et al.}~\cite{zhong2017toward} &  99.33 &  95.00  & 65.16$^\dagger$\\ 
&  ReST~\cite{wu2017recursive}    &  99.03 &  95.40 & - \\ 
&  GridFace~\cite{zhou2018gridface}   &  99.68 &  95.20 & -\\ 
& APA~\cite{wang2019adaptive}    &  99.68 &  - & - \\
&  Proposed  &  99.30 &  97.00 & 94.89 \\ 
&  Proposed + ArcFace Loss~\cite{deng2019arcface}   &  \textbf{99.78}  &  \textbf{97.90} & \textbf{98.03} \\ \hline
\end{tabular}
\end{center}
\end{table*}

\subsection{Visualization of Activation Maps by Grad-CAM}\label{sec:exp5}
To understand the effect of the proposed method qualitatively, we visualized the activation maps for input face images. 
As one of the visualization methods for visual explanations, we adopted the Grad-CAM~\cite{selvaraju2017grad} algorithm that uses the gradients of the target flowing into the final convolutional layer to show the localization map highlighting the important regions~\cite{selvaraju2017grad}. 
Note that the final convolution layer corresponds to the output of the feature map embedding layer ($\varphi$). 
Specifically, after computing the gradient of the score for the target for the embedded feature maps $\varphi\left(\mathbf{h}^{F}\right)$, the gradients flowing back were globally average-pooled to return the neuron importance weights. 
Then, the forward activation maps were combined based on the neural importance weights, then activated by the ReLU~\cite{selvaraju2017grad}. 
The images used for the visualization were obtained from a Web with the same ID label corresponding to the VGGFace2 dataset without face detection, which were not used for training. 

Fig.~\ref{fig_gradcam} shows the visualization results of the important feature by the Grad-CAM. 
As shown in the second and third rows, the activations without considering the face shape prior showed lower activation values and were sparsely spread. 
However, as shown in the fourth and fifth rows, the proposed method trained with the face shape prior concentrated on the important face regions (eyes, nose, and mouth). 
Less activation occurred in the excluded area that is not closely related face component. 
Also, the intensity values of the activation map obtained from the proposed method had relatively high values compared to those considering the face feature extraction network only. 

\subsection{Comparison to the State-of-the-art Methods}\label{sec:exp6}
Finally, we compared the proposed method to the state-of-the-art methods including both deep FR methods~\cite{liu2017sphereface, wang2018cosface, cao2018vggface2, deng2019arcface, duan2019uniformface, wei2020minimum, cao2020domain, huang2019deep, huang2020curricularface, zhang2021arface} and the face alignment learning algorithms~\cite{zhong2017toward, wu2017recursive, zhou2018gridface, an2019apa} for LFW, YTF, and MegaFace datasets.
We brought the FR performance from~\cite{deng2019arcface,duan2019uniformface,wei2020minimum,cao2020domain,huang2019deep,huang2020curricularface} (for deep FR methods) and~\cite{wei2020balanced} (for face alignment learning). 
In the case of VGGFace2, since the official performance for the MegaFace dataset is not reported in~\cite{deng2019arcface}, we measured the performance by using the publicly available VGGFace2 model.
For the proposed method, we additionally trained the network based on the ArcFace Loss~\cite{deng2019arcface} in order to accommodate the increase of the FR performance, which is denoted as ``Proposed + ArcFace Loss''.
The residual unit in the face feature extraction network (modified ResNet50) was replaced with the improved residual unit as used in~\cite{deng2019arcface}. 

As can be seen in Table~\ref{tab_sota}, the performance of the deep FR algorithms shows quite high performance, but its own face alignment algorithm is necessarily required. 
Since such a specific face alignment algorithm is needed, the performance can be degraded when the alignment type is changed or when the face detection or facial keypoint estimation algorithm fails in a wild environment. 
In addition, the facial keypoint estimation algorithm for the face alignment requires additional computations. 
However, the proposed method as a face alignment learning algorithm shows comparable performance without the specific face alignment. 
Therefore, the proposed method can be used for the FR system efficiently because it is not sensitive to the alignment type and does not require additional computations.
In addition, the proposed method outperforms the performance for the face alignment learning methods as shown in Table~\ref{tab_sota}.
Therefore, the proposed method would be preferable in terms of effectiveness and efficiency. 
In other words, the previous face alignment learning algorithms are designed with the face localization network (\textit{e.g.}, recursive spatial transformer modules in ReST~\cite{wu2017recursive}, rectification network and denoising autoencoder in GridFace~\cite{zhou2018gridface}) for a face alignment and the feature extraction network for FR, then trained in an end-to-end manner. 
In the testing phase, the input face image is forwarded to the localization network as well as the feature extraction network.
In contrast, since we can extract robust face features to the face misalignment from the face feature extraction network only, our model enables efficient inference in terms of computations and memories.

\subsection{Discussion}\label{sec:discussion}
In this paper, we focus on the observation that the performance of the existing deep face recognition algorithm is degraded if a face image is not well-aligned as used in training. Motivated by recent studies on face alignment learning that learns face alignment and face feature extraction in an end-to-end manner, we propose a face shape-guided face recognition algorithm based on feature alignment (\textit{i.e.}, pixel and feature alignments). Through the experiments with controlled and randomly aligned face images, we observed that the performance of the existing deep face recognition algorithm changed sensitively. In contrast, the proposed method showed robust recognition performance to the face misalignment. This result can be attributed to the proposed method not only being able to see various alignment types during training, but also finding face features by the devised feature alignments using the face shape as a clue (please refer to Fig.~\ref{fig_gradcam}). In addition, as shown in Table~\ref{tab_sota}, the proposed method showed comparable performance in a fair experimental environment that each deep face recognition algorithm performs optimal face alignment. Considering that the proposed method is an end-to-end framework integrating both face alignment and face feature extraction, it can be considered efficient. Moreover, we observed that the proposed method outperformed the existing face alignment learning algorithms. In the existing face alignment learning, a localization network in both training and testing is introduced. In the proposed method, the entire network with the help of the localization network (\textit{i.e.}, FAN) is trained via classification and feature alignment losses. However, the localization network is not required in testing. Therefore, our method is computationally efficient even compared to the existing face alignment learning algorithms.

\section{Conclusion} \label{sec:con}
In this paper, we proposed the face shape-guided deep feature alignment framework for FR robust to the face misalignment. 
Based on the face shape prior (\textit{i.e.}, face keypoints), we introduced two additional pixel alignment and feature alignment processes with the conventional face feature extraction network, which were learned in an end-to-end manner. 
For training, the proposed method learned the features for the well-aligned face image by decoding the aggregated features based on a face image and face shape prior.
In addition, through the feature alignment process, the learned feature was connected to align with the face shape-guided feature.
Through comparative experiments with LFW, CALFW, YTF, and MegaFace datasets, we validated the effectiveness of the proposed method toward face misalignment.
In particular, because we do not require additional computations for estimating the face keypoints and the face alignment, it would be efficient for testing a face image. 

\section*{Acknowledgment}
This work was supported by Institute of Information \& Communications Technology Planning \& Evaluation~(IITP) grant funded by the Korea government~(MSIT) (No.2014-3-00123, Development of High Performance Visual BigData Discovery Platform for Large-Scale Realtime Data Analysis and No.2020-0-00004, Development of Previsional Intelligence based on Long-term Visual Memory Network).

\bibliographystyle{IEEEtran}
\bibliography{bare_jrnl_compsoc}
\end{document}